\documentclass[preprint,review,3p,11pt,times]{elsarticle}

\usepackage{amssymb}
\usepackage{amsthm}
\usepackage{amsmath}
\usepackage{xcolor}
\usepackage{subfig}
\usepackage{lipsum}
\usepackage{pifont}
\usepackage{colortbl}
\usepackage{multirow}
\usepackage{comment}
\usepackage{ctable} % for \specialrule command
\biboptions{sort&compress}

\usepackage{algpseudocode}
\usepackage{amsfonts}
\usepackage[linesnumbered,ruled,vlined,longend]{algorithm2e}
\SetCommentSty{mycommfont}
\SetKwInput{KwInput}{Input}                % Set the Input
\SetKwInput{KwOutput}{Output}              % Set the Output

\usepackage{lineno}
\journal{Combustion and Flame}

\begin{document}

\begin{frontmatter}

\title{A Machine Learning Based Approach for Statistical Analysis of Detonation Cells from Soot Foils} 


\author[inst1]{Vansh Sharma\corref{cor1}}
\cortext[cor1]{Corresponding author.}
\ead{vanshs@umich.edu}

\author[inst1]{Michael Ullman}

\author[inst1]{Venkat Raman}
% \ead{ramanvr@umich.edu}

\affiliation[inst1]{organization={Department of Aerospace Engineering, University of Michigan},%Department and Organization
            city={Ann Arbor},
            postcode={MI 48109}, 
            country={USA}}

\begin{abstract}

This study presents a novel algorithm based on machine learning (ML) for the precise segmentation and measurement of detonation cells from soot foil images, addressing the limitations of manual and primitive edge detection methods prevalent in the field. Using advances in cellular biology segmentation models, the proposed algorithm is designed to accurately extract cellular patterns without a training procedure or dataset, which is a significant challenge in detonation research. The algorithm's performance was validated using a series of test cases that mimic experimental and numerical detonation studies. The results demonstrated consistent accuracy, with errors remaining within 10\%, even in complex cases. The algorithm effectively captured key cell metrics such as cell area and span, revealing trends across different soot foil samples with uniform to highly irregular cellular structures. Although the model proved robust, challenges remain in segmenting and analyzing highly complex or irregular cellular patterns. This work highlights the broad applicability and potential of the algorithm to advance the understanding of detonation wave dynamics.

\end{abstract}


\end{frontmatter}

%%%%%%%%%%%%%%%%%%%%%%%%%%%%%%%%%%%%%%%%%%%%%%%%%%%%%%%%%%%%%%%%%%%%%%%%%%%%%%%
%%%%%%%%%%%%%%%%%%%%%%%%%%%%%%%%%%%%%%%%%%%%%%%%%%%%%%%%%%%%%%%%%%%%%%%%%%%%%%%
\section{Introduction \label{sec:intro}}

Numerical and experimental studies of detonation wave propagation have played a key role in advancing a number of research areas and applications, including novel detonation-based combustors, \cite{raman2023nonidealities, Wolanski, Kailasanath, gutmarkpecs_rde}, safety measures and accident prevention protocols in industrial environments \cite{yanez2015analysis, yang2021review, ng2008_explosion}, and condensed phase explosives \cite{short_arfm_2018, voelkel2022effect}. Many such studies consider canonical configurations in order to isolate and explore critical aspects of detonation dynamics, including deflagration-to-detonation transition (DDT) length \cite{kuznetsov2005ddt}, reaction zone thickness, and detonation cell size \cite{crane2019isolating}. These parameters are vital for understanding the fundamental behavior and stability of detonation waves \cite{lee1984dynamic}. A common canonical configuration is a detonation propagating through a channel or tube. Here, soot foils are a widely used experimental technique for visualizing and measuring the cellular instabilities of detonations \cite{strehlow1968_cf, strehlow1974_aa, achasov2002dynamics, pintgen2003detonation, kellenberger2017simultaneous}. Soot foil measurements serve as a key point of comparison between experiments and simulations, while also aiding in the development of reduced-order models for detonation behavior \cite{crane2021geometric}. As such, it is critical for detonation researchers to be able to efficiently and accurately extract key measurements and morphological characteristics from experimental or numerical surrogate soot foils.

\begin{figure}[!hbt]
    \centering
    \includegraphics[clip,width=0.9\textwidth]{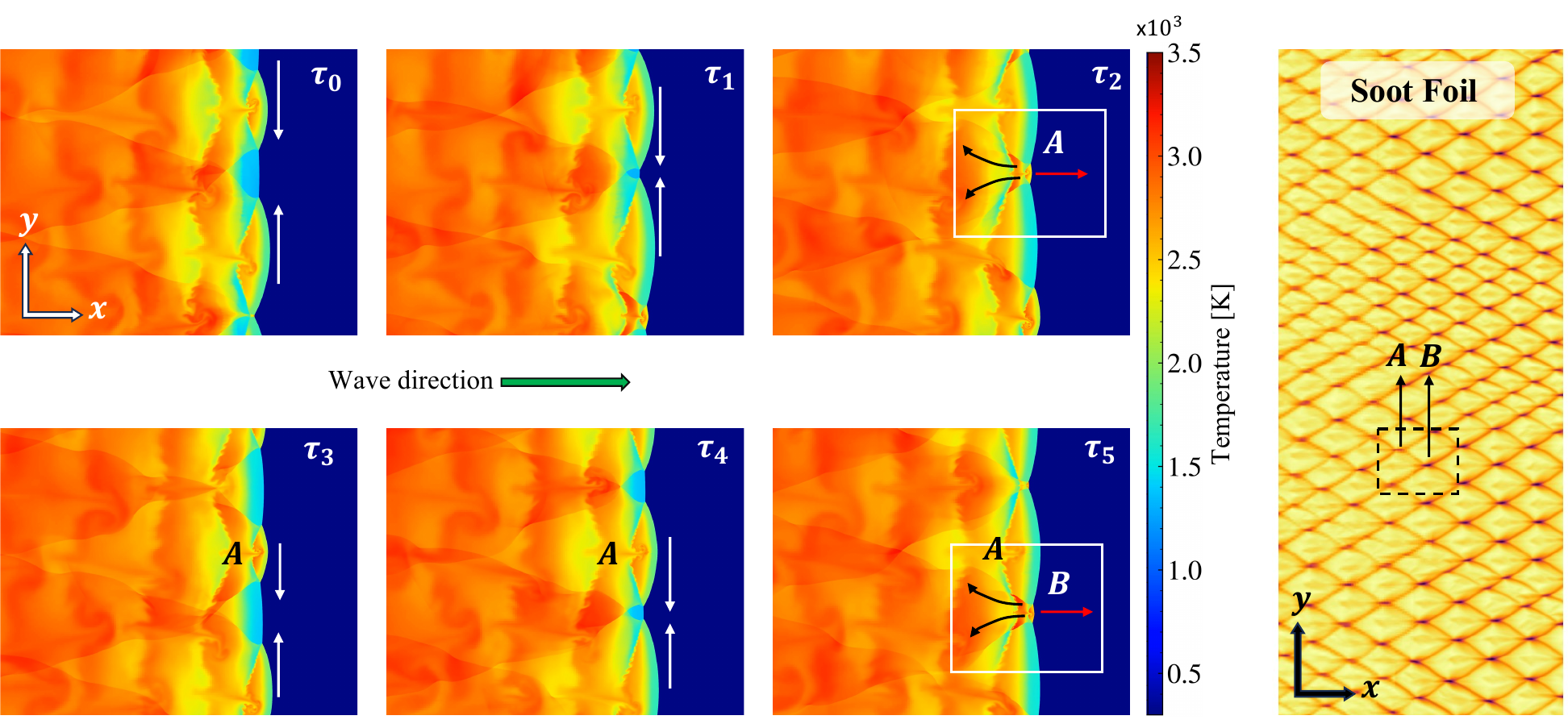}%
    \caption{(Left) Time sequence of temperature contours ($\tau_0$ to $\tau_5$) showing the evolution of a planar detonation wave front with triple point collisions at A and B. (Right) Numerical soot foil showing triple point trajectories through space.}
    \label{fig:sootfoil_seq}
\end{figure}

To illustrate the transient processes captured by soot foils, Fig.~\ref{fig:sootfoil_seq} presents a planar detonation wave propagating in a channel with premixed reactants. Six sequential snapshots ($\tau_0$ to $\tau_5$) of near-wave temperature fields highlight the instabilities along the detonation front as it propagates through the mixture. From $\tau_0$ to $\tau_2$, two transverse waves (indicated by the white arrows) move toward each other until they collide at point A at $\tau_2$. This collision creates a region of high pressure and temperature, which accelerates the resulting Mach stem ahead of the adjacent weaker regions, as highlighted in the white box. As the wave front progresses, the gases expand and the strength of the local shock wave diminishes until the next collision with an adjacent transverse wave at $\tau_5$. As such, the collision of the triple points at A gives rise to the next set of triple points that collide at B. Evidently, the detonation front exhibits coupled longitudinal-transverse instabilities, characterized by periodic oscillations of the leading shock and transverse waves through space and time. This gives rise to the cellular structure of detonation waves, where the trajectories of triple points (i.e., the intersection points between leading and transverse shock waves) form a diamond or ``fish-scale" pattern.

The width of the detonation cell ($\lambda$) is a key metric for assessing the ability of a mixture to stably detonate, known as its ``detonability" \cite{shepherd2009detonation, carter2022direct}. The size and regularity of the detonation cells depend on the chemical activation energies and the thermodynamic properties of the mixture, which vary with pressure, temperature, fuel type, and equivalence ratio \cite{lee_textbook, austin_thesis, jackson2013_cf}. Cell size is typically minimized near the stoichiometric fuel-air ratio and increases as the mixture becomes leaner or richer \cite{ciccarelli1994detonation, stamps1991hydrogen, vasil2006cell}. Experimentalists can record the cellular structure by coating the surface of the detonation tube or channel with soot. The high pressures of triple points leave imprints in the coating, allowing their trajectories to be recorded through space and time \cite{strehlow1968_cf, strehlow1974_aa, achasov2002dynamics}. These techniques are adaptable to various experimental setups and cost-effective compared to high-frequency Schlieren photography. Numerical soot foils, like the one shown on the right in Fig.~\ref{fig:sootfoil_seq}, replicate this process by storing the maximum pressure at every spatial location over the duration of the simulation \cite{alexei2023, prakash2019_fuelstrat}. Soot foil analysis offers insights into the effects of initial temperature, reactant dilution \cite{alexei2023, kumar1990detonation}, and reactant stratification on detonation behavior \cite{ullman-2024, prakash2019_fuelstrat}.

Several techniques have been developed to measure the properties of the cellular structure using soot foils. In the work of Shepherd et al.~\cite{shepherd1988analyses}, a cell size estimation tool was proposed using a power spectral density method combined with an edge detection technique. Here, periodicity in the cellular pattern was linked to the dominant frequencies using a 2D Fourier analysis. Nair et al.~\cite{nair2023detonation} utilized a 2D Fast Fourier Transform (FFT) to analyze wave behavior in an annular combustor. However, direct soot foil measurements were used to estimate the cell width, which was subsequently related to a characteristic length of the combustor. Using a directional gradient method with manual input, Carter et al.~\cite{carter2022direct} extracted cell lines and overlaid images to obtain an approximate structure of the detonation as a binary image. In the work of Ng et al.~\cite{ng2024detonation}, CH$^*$ chemiluminescence was used to extract the soot foil pattern; however, a manual method and a software package based on \cite{shepherd1988analyses} were used to perform cell size measurements. Meanwhile, Siatkowski et al.~\cite{siatkowski2021experimental} employed CAD software to plot lines in an image of a soot foil to extract cell sizes and developed an ML model to predict cell sizes \cite{siatkowski2024predicting}. Neural networks (NN) were used by Bakalis et al.~\cite{bakalis2023detonation} to predict the sizes of detonation cells; however, they relied on existing databases \cite{Shepherd-database} to train their network. Generally, the measurement of soot foils has entailed extensive manual input or the use of primitive edge detection methods that result in custom algorithms. As such, a generalizable approach that measures cell sizes with minimal user input is still lacking.

Based on the aforementioned challenges, the objective of the current work is to present an algorithm based on a machine learning (ML) model that can perform precise image segmentation and accurate measurements of detonation cells. Image segmentation and pattern recognition are central to computer vision \cite{suri-2000}, and the soot foil measurement problem can be reframed in this context. Labeled images of soot foils with clearly marked patterns are essential for models to accurately extract the cell size. However, the detonation community faces a significant problem due to the scarcity of training data required for ML models. An interesting parallel exists in the field of quantitative cellular biology, where diverse sets of cell images collected from different tissues are used to identify key cellular properties such as shape, position, and RNA expression. Stringer et al.~\cite{cellpose_stringer-2020} developed a generalized model for image segmentation within this context. Given the morphological similarities between biological cells and the patterns seen in soot foils, this approach is particularly suitable to address the current problem. The current study focuses on developing a robust algorithm —based on the work of Stringer et al.—capable of measuring detonation cell properties from soot foils and providing insight into the underlying wave dynamics driving different cell morphologies. The efficacy of the algorithm is evaluated in various scenarios, highlighting its potential to be widely and easily applied in detonation research.

%%%%%%%%%%%%%%%%%%%%%%%%%%%%%%%%%%%%%%%%%%%%%%%%%%%%%%%%%%%%%%%%%%%%%%%%%%%%%%%
%%%%%%%%%%%%%%%%%%%%%%%%%%%%%%%%%%%%%%%%%%%%%%%%%%%%%%%%%%%%%%%%%%%%%%%%%%%%%%%
\section{Cell measurement algorithm \label{sec:MLAlgo}}
The measurement algorithm\footnote{\text{\color{red} Code to be shared after the paper is accepted.}} consists of three main steps: preprocessing, image segmentation, and feature extraction. These are detailed in the following subsections.

\begin{comment}
\begin{enumerate}
    \item \textit{Preprocessing}: De-noising the image to remove marks and noise and enhancing the cell edges using edge-detection gradient based algorithms.
    \item \textit{Image segmentation}: Applying the image segmentation ML model.
    \item \textit{Feature extraction}: Computing cell properties and other quantities of interest from the feature masks.
\end{enumerate}
\end{comment}

%%%%%%%%%%%%%%%%%%%%%%%%%%%%%%%%%%%%%%%%%%%%%%
%%%%%%%%%%%%%%%%%%%%%%%%%%%%%%%%%%%%%%%%%%%%%%
%%%%%%%%%%%%%%%%%%%%%%%%%%%%%%%%%%%%%%%%%%%%%%
%%%%%%%%%%%%%%%%%%%%%%%%%%%%%%%%%%%%%%%%%%%%%%

\subsection{Preprocessing \label{sec:preProc}}
Prior to applying the ML model, the input image undergoes a series of steps aimed at improving the accuracy of the subsequent segmentation. These steps include:

\begin{itemize}
    \item \textbf{De-noising}: The raw image $I(x,y)$ is smoothed using a Gaussian filter $G(x, y)$ to reduce noise while preserving essential features such as edges:
    \begin{equation}
        I_s(x, y) = G(x, y) \circledast I(x, y)
        \label{eqn:denoise}
    \end{equation}
    Here, $\circledast$ is the convolution operator. This step helps to remove unwanted artifacts from the image, which can otherwise lead to inaccurate segmentation results. Furthermore, denoising can be performed using the built-in denoising utility of~\cite{cellpose_stringer-2020}. However, oversmoothing can blur important features, which also negatively affects the segmentation.
    
    \item \textbf{Edge Detection with Directional Bias}: The smoothed image $I_s(x, y)$ is processed to emphasize edges that run in specific directions (e.g. north-south). Gradients are calculated in the $x$- and $y$-directions by
    \begin{equation}
        \begin{split}
            G_x(x, y) &= I_s(x+1, y) - I_s(x-1, y) \\
            G_y(x, y) &= I_s(x, y+1) - I_s(x, y-1)
        \end{split}
        \label{eqn:grad_dir_bias}
    \end{equation}
    These are then combined with coefficients $\alpha$ and $\beta$ to enhance edges in the desired orientation:
    \begin{equation}
        G_{\text{bias}}(x, y) = \alpha G_y(x, y) + \beta G_x(x, y)
        \label{eqn:grad_bias}
    \end{equation}
    This directional bias can help highlight specific features, but it may also obscure or downplay edges that are not aligned with the chosen direction.
    
    \item \textbf{Non-Maximum Suppression}: To reduce the thicknesses of edges, non-maximum suppression is applied. This results in a binary edge map:
    \begin{equation}
        E(x, y) = 
        \begin{cases} 
        G_{\text{bias}}(x, y) & \text{if } G_{\text{bias}}(x, y) \text{ is a local maximum along the gradient direction} \\
        0 & \text{otherwise}
        \end{cases}
        \label{eqn:edge_map}
    \end{equation}
    This step ensures that only the most prominent edges are retained, which improves the clarity of segmentation. However, it can also result in the loss of weaker edges that might be significant in some contexts.
    
    \item \textbf{Edge Enhancement by Overlay}: The edge map is enhanced by overlaying the image multiple times:
    \begin{equation}
        E_{\text{enhanced}}(x, y) = \sum_{n=1}^{N} E(x, y)
        \label{eqn:edge_enhance}
    \end{equation}
    This process amplifies the detected edges, making them more prominent for the segmentation step. A potential downside here is that this could exaggerate noise if not carefully controlled.
\end{itemize}
These preprocessing steps ensure that the edges of the cells are well defined and that noise is minimized, thereby improving the performance of the subsequent image segmentation step. It is important to note that these preprocessing steps are optional and may be applied depending on the quality and noise level of the image. This is discussed further in \ref{appendix:preProc}.

\subsection{Segmentation model \label{sec:seg_model}}

\begin{figure}[!hbt]
    \centering
    \includegraphics[clip,width=0.98\textwidth]{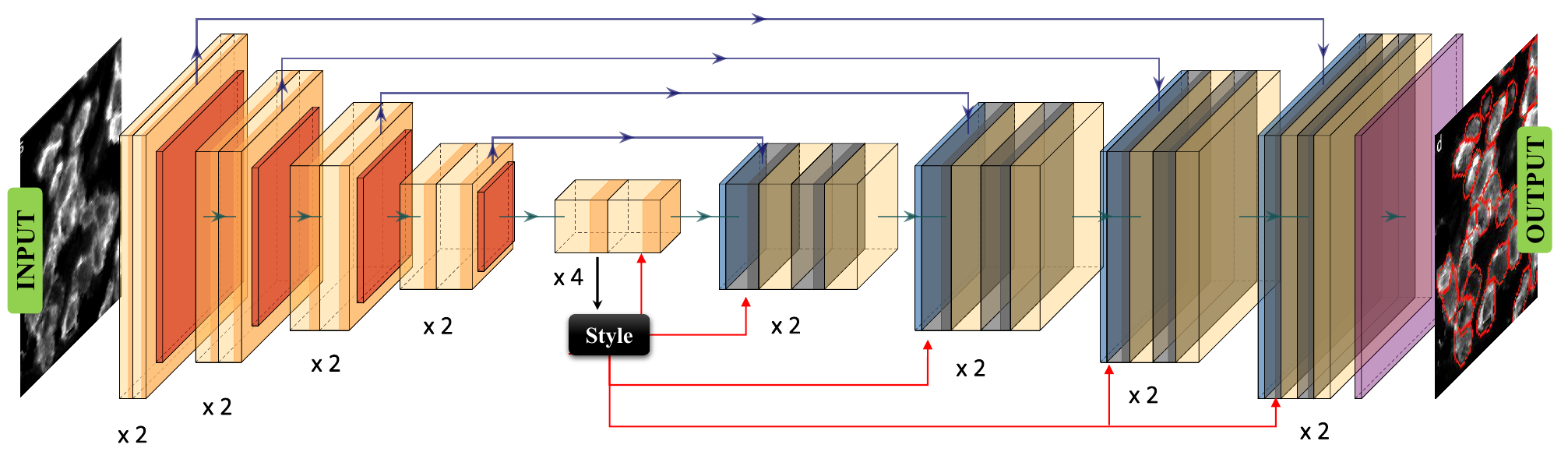}%
    \caption{UNet architecture for cell segmentation task adapted from \cite{cellpose_stringer-2020}}.
    \label{fig:unet_arch}
\end{figure}

The ML model \cite{cellpose_stringer-2020} uses a modified U-Net architecture \cite{UNet}, which downsamples and then upsamples convolutional maps in a mirror-symmetric manner. This is illustrated in Fig.~\ref{fig:unet_arch}. Instead of feature concatenation, direct summation integrates convolutional maps from the downsampling path with those in the upsampling path, thereby reducing the number of parameters. Standard U-Net blocks are replaced with residual blocks for improved performance, and the network depth is doubled. In addition, global average pooling is applied to the smallest convolutional maps to extract a ``style" vector \cite{styletransfer}, defined as the global average pool of each feature map. This style vector is incorporated into all upsampling stages to account for image-specific processing variations. For further details on the model architecture, the reader is referred to \cite{cellpose_stringer-2020}. Different variations of the aforementioned ML model were trained using a cycle generative adversarial network approach (cGAN-Seg) \cite{GANcellpose_zargari-2024} for scenarios with limited datasets. However, the fully pre-trained model is used here due to its higher accuracy. As such, there are three available pre-trained models—\textit{cyto} (hereafter referred to as \textit{cyto1}), \textit{cyto2}, and \textit{cyto3}. We first demonstrate the capabilities of \textit{cyto1} and \textit{cyto2}, which are designed for image segmentation and trained on progressively larger datasets. The \textit{cyto3} model \cite{Cellpose3} is trained on the largest dataset and is capable of processing noisy images without any preprocessing. The enhanced edge map $E_{\text{enhanced}}(x, y)$ is input into the machine learning model $f_{\text{ML}}$ to obtain the segmentation mask:
\begin{equation}
    M(x, y) = f_{\text{ML}}(E_{\text{enhanced}}(x, y))
    \label{eqn:mask}
\end{equation}

%%%%%%%%%%%%%%%%%%%%%%%%%%%%%%%%%%%%%%%%%%%%%%
%%%%%%%%%%%%%%%%%%%%%%%%%%%%%%%%%%%%%%%%%%%%%%
%%%%%%%%%%%%%%%%%%%%%%%%%%%%%%%%%%%%%%%%%%%%%%
%%%%%%%%%%%%%%%%%%%%%%%%%%%%%%%%%%%%%%%%%%%%%%

\subsection{Feature measurement \label{sec:feature_measurement}}
The input image is evaluated by the model, which yields a label mask $M(x, y)$ where each pixel $(x, y)$ is assigned a label corresponding to a specific segmented cell. The generated masks represent distinct cell boundaries within the image, from which an outline is computed. The visualization of these outlines overlaid on the original image is done to confirm successful cell segmentation. The statistics for cell segmentation are computed using the following methods. Translation to physical quantities is performed simply by scaling with the physical dimensions of the image.
 
\begin{itemize}

    \item\textbf{Area Calculation:} The area for each unique cell label $k$ (ignoring the background label 0) is calculated by summing over the number of pixels associated with each cell in the mask. This pixel count is translated directly into cell areas in pixels, providing a quantitative measure of cell size:
    \begin{equation}
        A_k = \sum_{(x, y) \in M(x, y)} \delta(M(x, y) - k)
        \label{eqn:cell_area}
    \end{equation}
    where $\delta(M(x, y) - k)$ is an indicator function that is 1 if $M(x, y) = k$ (i.e., the pixel belongs to cell $k$) and 0 otherwise.
    
    \item\textbf{Centroid Calculation:} The centroid of each cell is calculated using the center-of-mass formula, which provides the coordinates of the geometric center of the cell:
    \begin{equation}
        C_x^k = \frac{\sum_{(x, y) \in M(x, y)} x \cdot \delta(M(x, y) - k)}{A_k}, \quad C_y^k = \frac{\sum_{(x, y) \in M(x, y)} y \cdot \delta(M(x, y) - k)}{A_k}
        \label{eqn:cell_centroid}
    \end{equation}
    where $C_x^k$ and $C_y^k$ are the $x$- and $y$-coordinates of the centroid of the $k$-th cell, respectively.
    
    \item\textbf{Bounding Box Calculation:} The bounding box of each cell is computed to determine its spatial extent as follows:
    \begin{equation}
        (i,j)_{X\min}^k = \min_{(x, y) \in M(x, y)} \{x_{dir} \mid M(x, y) = k\}, \quad (i,j)_{X\max}^k = \max_{(x, y) \in M(x, y)} \{x_{dir} \mid M(x, y) = k\}
        \label{eqn:bounding_box_x}
    \end{equation}
    \begin{equation}
        (i,j)_{Y\min}^k = \min_{(x, y) \in M(x, y)} \{y_{dir} \mid M(x, y) = k\}, \quad (i,j)_{Y\max}^k = \max_{(x, y) \in M(x, y)} \{y_{dir} \mid M(x, y) = k\}
        \label{eqn:bounding_box_y}
    \end{equation}
    
    The size of the region in the $x$- and $y$-directions is then given by:
    \begin{equation}
        \begin{split}
            \text{Size}_x^k &= x_{\max}^k - x_{\min}^k + 1 \\
            \text{Size}_y^k &= y_{\max}^k - y_{\min}^k + 1
        \end{split}
        \label{eqn:box_size}
    \end{equation}
    
    \item\textbf{Major and Minor Axes Calculation:} The major and minor axes are inferred from the span of the cell label indices using the Euclidean distance. These measurements are averaged for a subset of sampled cells to estimate typical cell extents:
    \begin{equation}
        \begin{split}
            \text{MajorAxis}_k &= (i,j)_{X\max}^k \boldsymbol{\cdot} (i,j)_{X\min}^k \\
            \text{MinorAxis}_k &= (i,j)_{Y\max}^k \boldsymbol{\cdot} (i,j)_{Y\min}^k
            \label{eqn:box_axes}
        \end{split}
    \end{equation}
\end{itemize}

The complexity involved in these calculations stems from the need to 1) accurately map mask pixels to specific cells, 2) ensure precise centroid determination, and 3) manage large image datasets efficiently. This computational rigor is ensured through various verification steps, such as confirming the reproducibility in random sampling of cell labels for further analyses of cell dimension statistics.

%%%%%%%%%%%%%%%%%%%%%%%%%%%%%%%%%%%%%%%%%%%%%%%%%%%%%%%%%%%%%%%%%%%%%%%%%%%%%%%
%%%%%%%%%%%%%%%%%%%%%%%%%%%%%%%%%%%%%%%%%%%%%%%%%%%%%%%%%%%%%%%%%%%%%%%%%%%%%%%
\section{Results and discussion \label{sec:results}}

The proposed algorithm was tested in multiple relevant scenarios to assess its limits on accuracy and precision. In addition, the cell measurement algorithm is adaptable to various computing hardware. The analysis was conducted on an Apple M2 Max with a 12-core CPU, 38-core GPU, 16-core Neural Engine, and 96GB RAM, focusing on CPU performance by omitting Metal Framework's GPU capabilities. The results were also reproduced using single and multiple NVIDIA H100 80GB HBM3 GPUs, demonstrating the adaptability of the framework in different computational environments. Note that larger input data arrays require more GPU VRAM for processing, but specific metrics are not provided as they are not of primary focus here. 

\subsection{Generated data \label{sec:generated_data}}

\begin{figure}[!h]
    \centering
    \includegraphics[clip,width=0.78\textwidth]{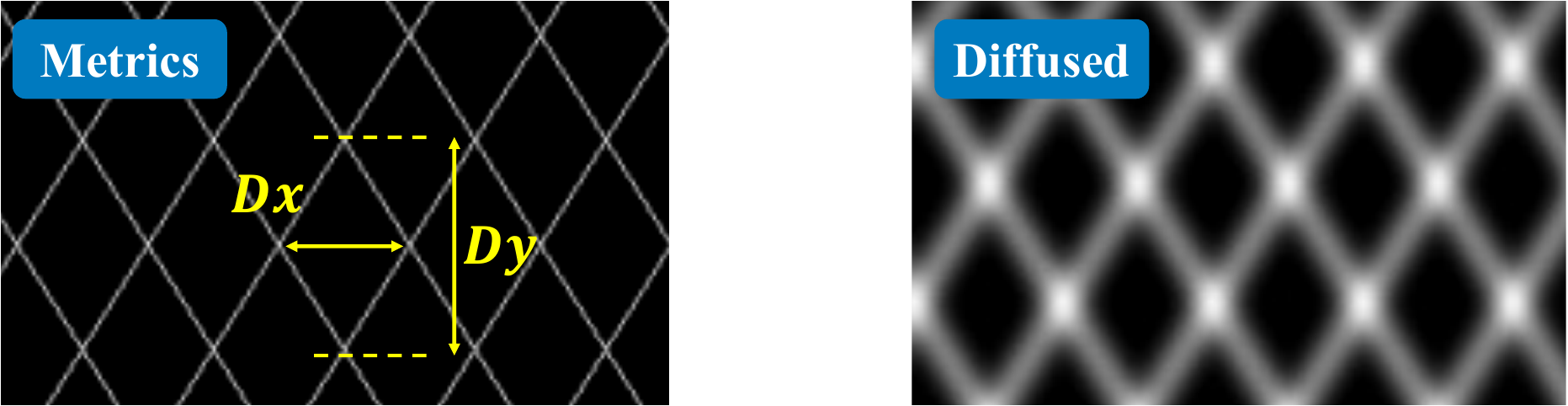}%
    \caption{Manufactured data shown on the left with metrics marked in yellow and the right plot shows the diffused image.}
    \label{fig:test_cell}
\end{figure}

The pretrained ML models, \textit{cyto1} and \textit{cyto2}, are first verified using artificially generated data to estimate their capabilities. The \textit{cyto3} model, which can be applied directly to noisy images, will be discussed in later sections. A rhombus-patterned set of soot foil data is generated to evaluate the applicability and performance of the models. The primary control parameters are the lengths of the major and minor axes, $Dy$ and $Dx$, as shown in Fig.~\ref{fig:test_cell}. The quantity $Dx$ represents the Euclidean distance between the two extreme points of the segmented cell along the $x$-direction, while $Dy$ is defined similarly for the $y$-direction. In addition, Gaussian noise is applied to the pattern, with the standard deviation of the noise distribution ($\sigma$) used as another control variable for testing. The resulting image, shown on the right in Fig.~\ref{fig:test_cell}, shows a soot foil pattern similar to those observed in both the experimental and numerical detonation studies. Here, a successful segmentation is defined as predicting cell dimensions within a 10\% error margin. In contrast, segmentation is considered to have failed if the error exceeds this threshold or if an insufficient number of cells are detected for meaningful analysis. 

\begin{table}[!hbt]
    \centering
    \begin{tabular}{|c|c|c|c|c|c|c|c|}
        \specialrule{.2em}{.1em}{.1em}
        \multirow{2}{*}{\textbf{Case}} & \multirow{2}{*}{\textbf{\textbf{$\boldsymbol{Dy}$}} }& \multirow{2}{*}{\textbf{\textbf{$\boldsymbol{Dx}$}}} & \multirow{2}{*}{\textbf{Filter $\boldsymbol{\sigma}$}} & \multirow{2}{*}{\textbf{Predicted $\boldsymbol{Dy}$}} & \multirow{2}{*}{\textbf{Predicted $\boldsymbol{Dx}$}} & \multicolumn{2}{c|}{\textbf{Model}} \\
        \cline{7-8}
        & & & & & & \cellcolor{blue!10}\textit{Cyto 1} & \cellcolor{blue!30}\textit{Cyto 2 / 3} \\
        \specialrule{.2em}{.1em}{.1em}
        1 & 50 & 50 & 1.5 & 47.9 $^{c2}$ & 48.4 $^{c2}$ & \cellcolor{green!30}\ding{51} & \cellcolor{green!30}\ding{51} \\
        \hline
        2 & 50 & 120 & 2.0 & 49.6 $^{c1}$ & 111.1 $^{c1}$ & \cellcolor{green!30}\ding{51} & \cellcolor{green!30}\ding{51} \\
        \hline
        3 & 120 & 50 & 1.5 & 108.8 & 49.0 & \ding{55} & \cellcolor{green!30}\ding{51} \\
        \hline
        4 & 120 & 10 & 2.0 & - & - & \cellcolor{red!30}\ding{55} & \cellcolor{red!30}\ding{55} \\
        \hline
        5 & 120 & 50 & 2.0 & 108.4 & 49.0 & \ding{55} & \cellcolor{green!30}\ding{51} \\
        \specialrule{.2em}{.1em}{.1em}
    \end{tabular}
    \caption{Generated data results with different segmentation models. Superscript (c1 and c2) used for model with higher accuracy.}
    \label{tab:test_case_results}
\end{table}

\begin{figure}[!hbt]
    \centering
    \includegraphics[clip,width=0.68\textwidth]{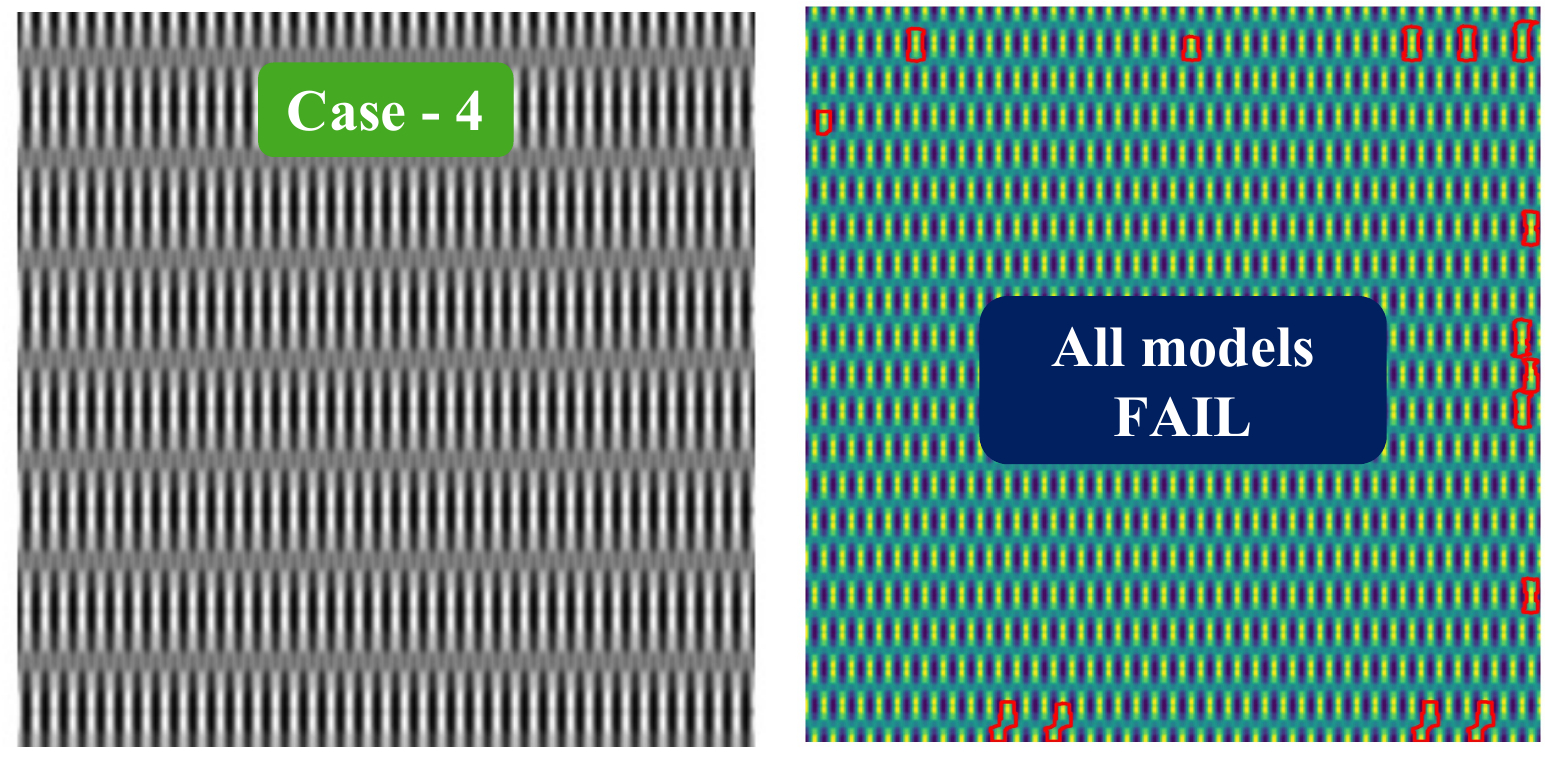}%
    \caption{Case 4 cellular pattern shown in the left plot and cells detected (red) by \textit{cyto2} model in the right plot.}
    \label{fig:case4}
\end{figure}

Table \ref{tab:test_case_results} presents the five test cases designed to examine the directional geometric skewness and noise tolerance of the models. The predicted variables were averaged across 10 random samples in the predicted image. The \textit{cyto1} model successfully segmented cells in cases 1 and 2, but failed in subsequent cases, where the models were unable to segment any cell. This failure is attributed to \textit{cyto1} having fewer training data and less diverse image sets compared to the latest models. The \textit{cyto2} model successfully segmented all cases except case 4. Figure \ref{fig:case4} illustrates the patterns and results for case 4 using the \textit{cyto2} model. The extreme skewness in case 4 posed significant challenges for both models. For cases where both models succeeded, the results of the superior model are reported, as indicated by superscripts in Table \ref{tab:test_case_results}. The maximum error across all cases is below 10\%. The results of the \textit{cyto3} model are excluded, as its performance is comparable to that of the \textit{cyto2} model for the given task.

A notable observation is that when the skewness is biased in the $y$-direction (i.e., along the major axis $Dy$), the accuracy of the model decreases. This often leads to the complete failure of the \textit{cyto1} model. There is no inherent restriction preventing the neural network from predicting the horizontal and vertical spans of cells that do not correspond to the realistic cell shapes used in the training data. Generally, the predicted cell span aligns with realistic cell shapes because the network was trained to have a consistent cell span. However, uncertainty in network predictions can lead to inconsistencies. During model training, the consistency of the predicted cell shapes is verified against the training data by computing the mean squared error between the span gradients (gradients in $x$- and $y$- directions) of the predictions and the training data \cite{cellpose_stringer-2020}. The inference parameters of the model can be tuned to capture skewed cellular shapes, as shown in the next section. 

\subsection{Practical data \label{sec:practical_data}}

When processing soot foil data from experiments or simulations, the primary segmentation challenges are 1) the quality of the soot foil image and 2) the irregularity of the detonation cells. These are respectively analogous to the previous noise and skewness tests with generated data. However, actual detonation cells exhibit additional irregularities, including spatial variations in orientation, area, and relative lengths of the major and minor axes. Noting this variability, the efficacy of the models is demonstrated using soot foil data from several studies with varying levels of irregularity in the detonation cells.

\subsubsection{Cell detection \label{sec:cell_detection}}

\begin{figure}[!h]
    \centering
    \includegraphics[clip,width=0.95\textwidth]{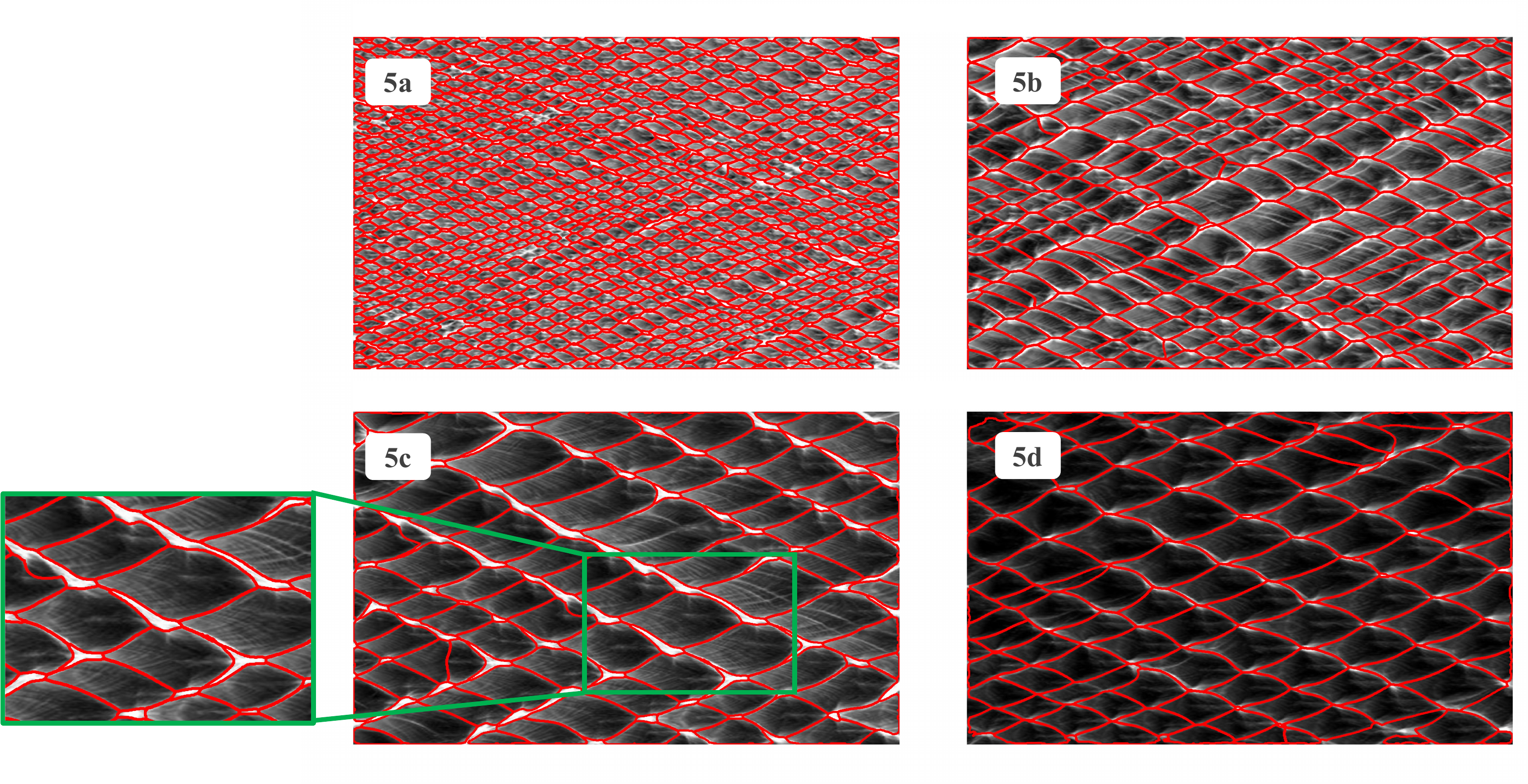}%
    \caption{Detected cells marked with red outlines overlaid on figures from Fig.\ 5 in \cite{smirnov2024modelling}.}
    \label{fig:smi_f5}
\end{figure}

The performance of the algorithm is tested against both the numerical and experimental soot foil data. Data from multiple studies \cite{smirnov2024modelling, carter2022direct, li2021influences, sugiyama2011characteristics} are utilized and soot foil images are extracted using the framework in \cite{vanshRAG}. In Fig.~\ref{fig:smi_f5}, the red outlines indicating the detected cellular boundary are overlaid on the grayscale images extracted directly from the article by Smirnov et al. \cite{smirnov2024modelling}. The quality of detection can be evaluated by considering the clarity of the cell boundaries, the accuracy of the outlines, and the level of detail captured in the image (see inset for 5c). In Fig.~\ref{fig:smi_f5}, the red outlines consistently mark the cell boundaries across all four images, indicative of the precision and reliability of the algorithm. This is notable in images 5a and 5b, where the cells are, respectively, tightly packed and irregularly shaped. However, it is essential to consider the potential for false positives or false negatives in the detection process, as well as the potential for variations in detection performance due to different experimental conditions or imaging parameters. Therefore, human supervision is recommended to ensure adequate performance.

\begin{figure}[!h]
    \centering
    \includegraphics[clip,width=0.95\textwidth]{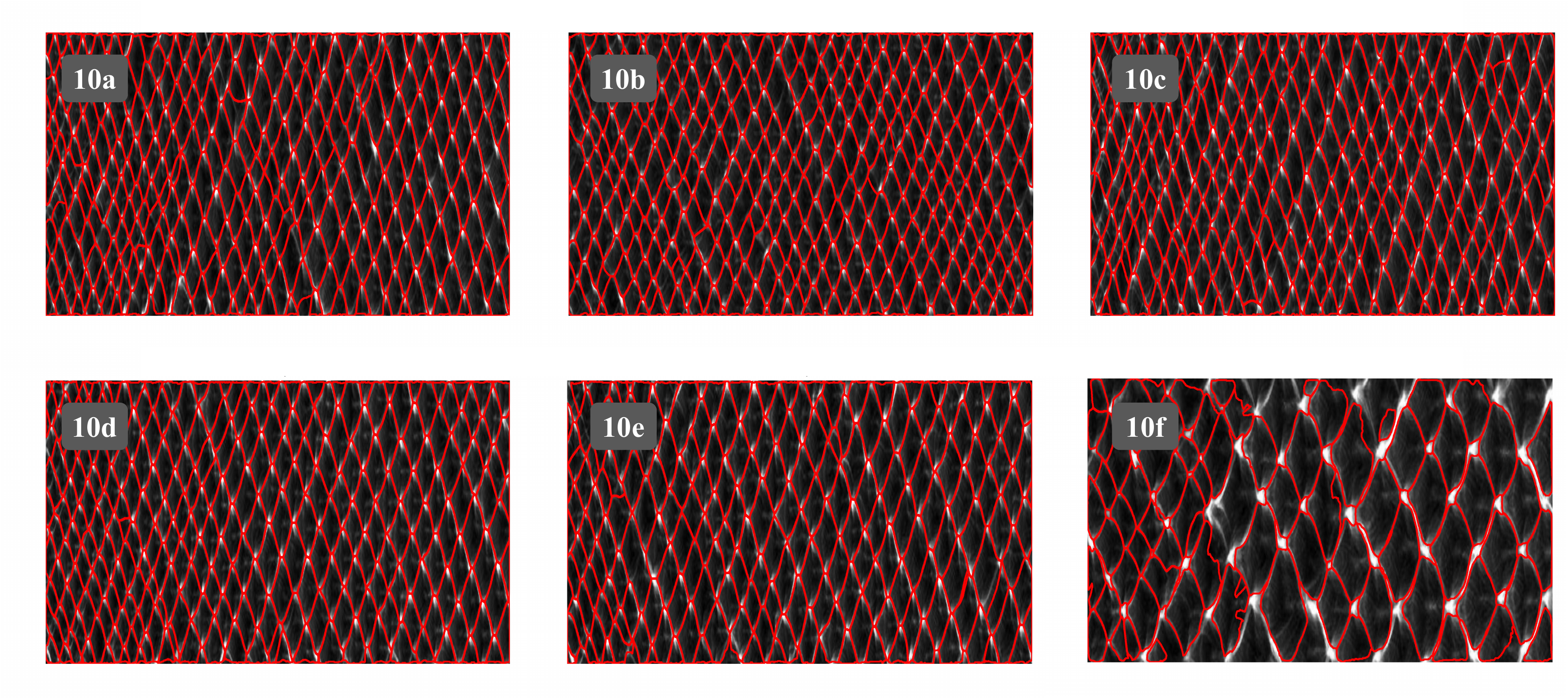}%
    \caption{Detected cells marked with red outlines overlaid on figures from Fig.\ 10 in \cite{smirnov2024modelling}.}
    \label{fig:smi_f10}
\end{figure}

In Fig.~\ref{fig:smi_f10}, six additional triple-point trajectory maps from Ref.~\cite{smirnov2024modelling} are analyzed. For plots 10a to 10c in their paper, the algorithm is able to precisely capture the distinct cellular pattern. Here, the \textit{cyto3} model was used for image segmentation with pre-proccessed images. In plot 10f, the cellular patterns near the upper and lower boundaries are poorly detected. Various models were tested, but successful detection was probably hindered by incomplete cell shapes. Although the boundary cells are similar in size to the internal cells, they are clipped into nearly half-shaped shapes, resulting in open rather than closed forms. Open shapes appear in other plots as well, but they are relatively small compared to the overall image dimensions. However, in plot 10f, the sizes of these open shapes are more significant.

\begin{figure}[!hbt]
    \centering
    \includegraphics[clip,width=0.95\textwidth]{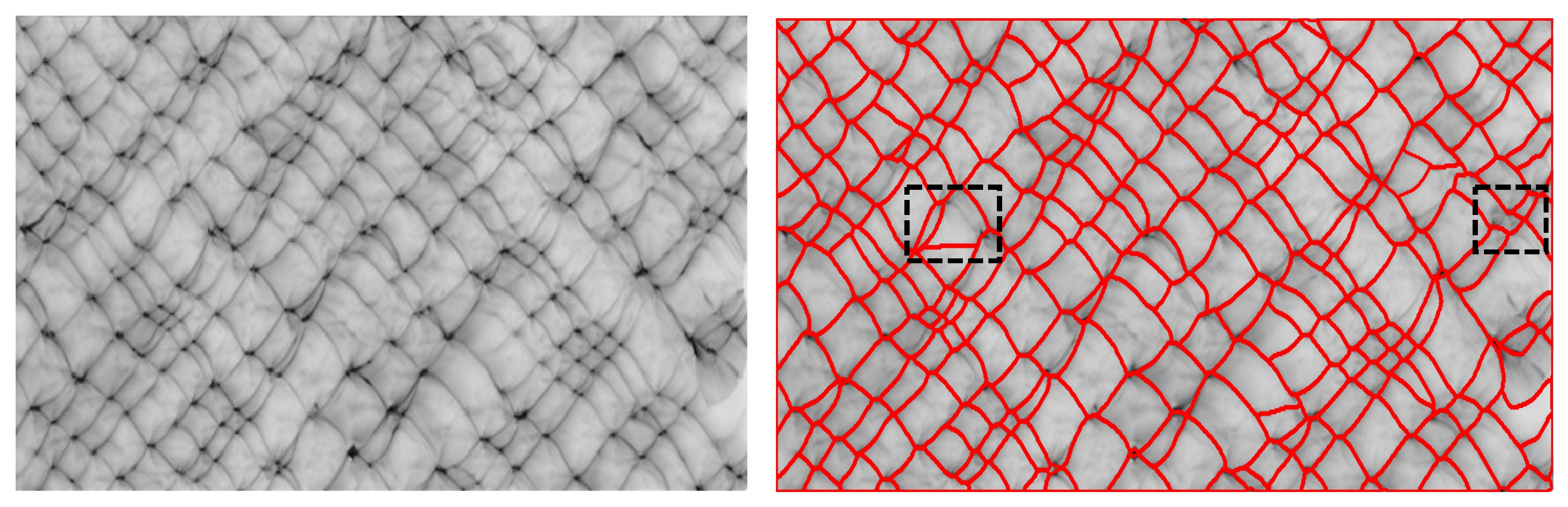}%
    \caption{Detected cells marked with red outlines overlaid on irregular soot foil from Fig.\ 3 in \cite{li2021influences}.}
    \label{fig:irreg}
\end{figure}

Figures \ref{fig:smi_f5} and \ref{fig:smi_f10} represent uniform cellular patterns, characterized by cells with consistent shapes, sizes, and orientations. In contrast, the left plot in Fig.~\ref{fig:irreg} shows a soot foil formed by irregular cellular detonations. This leads to noticeable variability in cell shapes, orientations, and spatial densities. The right plot in Fig.~\ref{fig:irreg} shows the results of cell detection using the \textit{cyto2} model. In this case, the raw image was directly analyzed, and additional segmentation iterations were used within the model to accurately capture the diverse and irregular cell shapes. Qualitatively, the segmentation task yields good results comparable to those of uniform cellular patterns. However, there are a few instances (black boxes in the right image) where cells were merged during detection, and, in one specific case, the edge was misaligned. Additional soot foil segmentation data is provided in~\ref{appendix:extraData}. A more quantitative assessment follows in Sec.\ \ref{sec:cell_metrics}, where statistics of quantities of interest (QOIs), such as cell area and axial dimensions, are calculated to provide a detailed analysis of segmentation accuracy.

\subsubsection{Cell metrics and statistics \label{sec:cell_metrics}}

\begin{comment}
For line 1 in Fig.~\ref{fig:f10c_pdf}, the statistics for cell area, $Dx$, and $Dy$ indicate that the cells are small and uniformly spaced. Namely, the cell area distribution is skewed with a peak in smaller areas, and the $Dx$ and $Dy$ distributions have relatively distinct peaks that indicate consistent horizontal and vertical spacing, respectively. Along line 2, the cells vary more in size and spacing, which is reflected in the wider distributions for cell area and $Dy$. Similarly for line 3, the wide and bimodal distributions for $Dx$ and $Dy$ indicate an irregularity in cell size, which is reflected in the bimodal distribution for cell area. On the other hand, for line 4, the distributions are more symmetric about their means, especially for $Dy$. The $Dx$ distribution for line 4 is wider than that for line 1, but narrower than those for lines 2 and 3. In general, the peaks of the $Dx$ distributions are located roughly at the same $Dx$ values for all four lines.

\begin{figure}[!h]
     \centering
     \includegraphics[clip,width=0.95\textwidth]{figures/Fig10_C_pdfs.pdf}%
     \caption{Statistical analysis of cell characteristics for Fig.\ 10c in \cite{smirnov2024modelling}. \textcolor{blue}{MU: Can we have the density and cell area axes have the same bounds? Otherwise, it is difficult to compare between them.}}
     \label{fig:f10c_pdf}
\end{figure}
\end{comment}

\begin{figure}[!h]
    \centering
    \includegraphics[width=0.85\textwidth]{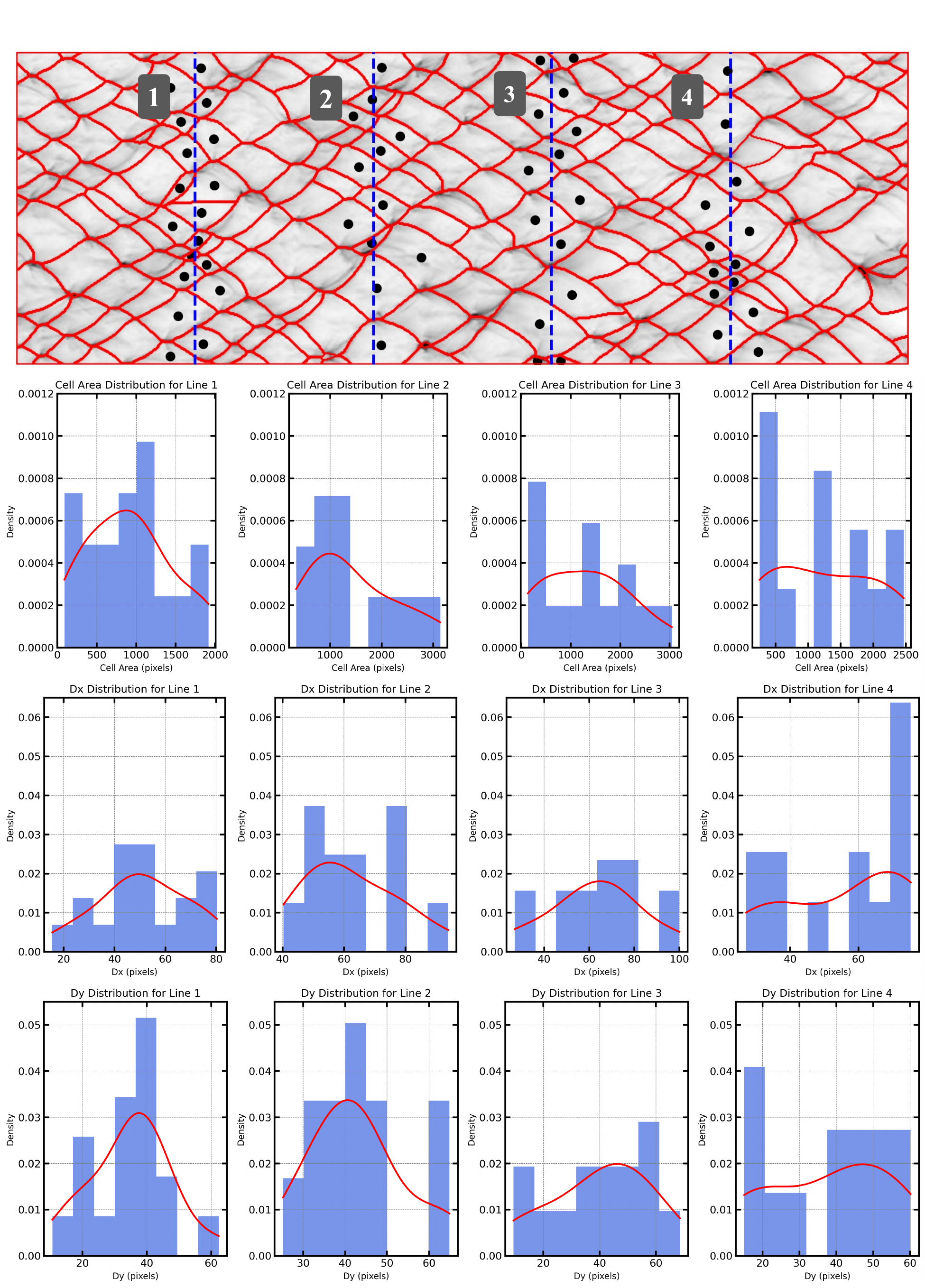}%
    \caption{Statistical analysis of cell characteristics for Fig.\ 3 in \cite{li2021influences}.}
    \label{fig:f3_pdf}
\end{figure}

Figure \ref{fig:f3_pdf} presents statistics for various cell metrics. The top row displays the segmented image from Fig.~\ref{fig:irreg}, with four vertical lines spaced equidistantly along the wave normal direction. The cells intersected by these lines are marked with black circles at their centroids. The second row of Fig.~\ref{fig:f3_pdf} shows four histograms of the cell areas along the corresponding lines. The third and fourth rows also present histograms for the same four lines, but for $Dx$ and $Dy$, respectively. The optimal bin count for the histogram is determined using a Bayesian approach described in \cite{knuth2006optimal}. Gaussian kernel density estimations (KDEs) are plotted in red for each of the histograms to visualize the distributions of the quantities observed in the intersected cells.

These statistics reflect the variability in cellular structure observed over the channel length. Notable trends can be seen in all three quantities from left to right. For instance, the cell area tends to be smaller and more consistent along line 1, but there is greater variability along the remaining lines as the distributions become wider and larger cells appear more frequently. This trend is also evident for the $Dx$ and $Dy$ distributions, which have longer tails for lines 2, 3, and 4. Indeed, along line 4, the KDEs for all quantities are roughly uniform with multiple small peaks—indicative of significant irregularities. Overall, these histograms illustrate how the cellular structure becomes progressively more heterogeneous along the length of the channel. Considering that the evolution of the cellular structure in simulations can depend on the initial conditions, analyzing the cellular patterns in this manner can be a useful tool for assessing simulation convergence. Namely, if the statistics for the cell sizes remain unchanged after a given distance, the detonation may be considered to be propagating in a quasi-steady-state.

\begin{figure}[!h]
    \centering
    \includegraphics[width=0.85\textwidth]{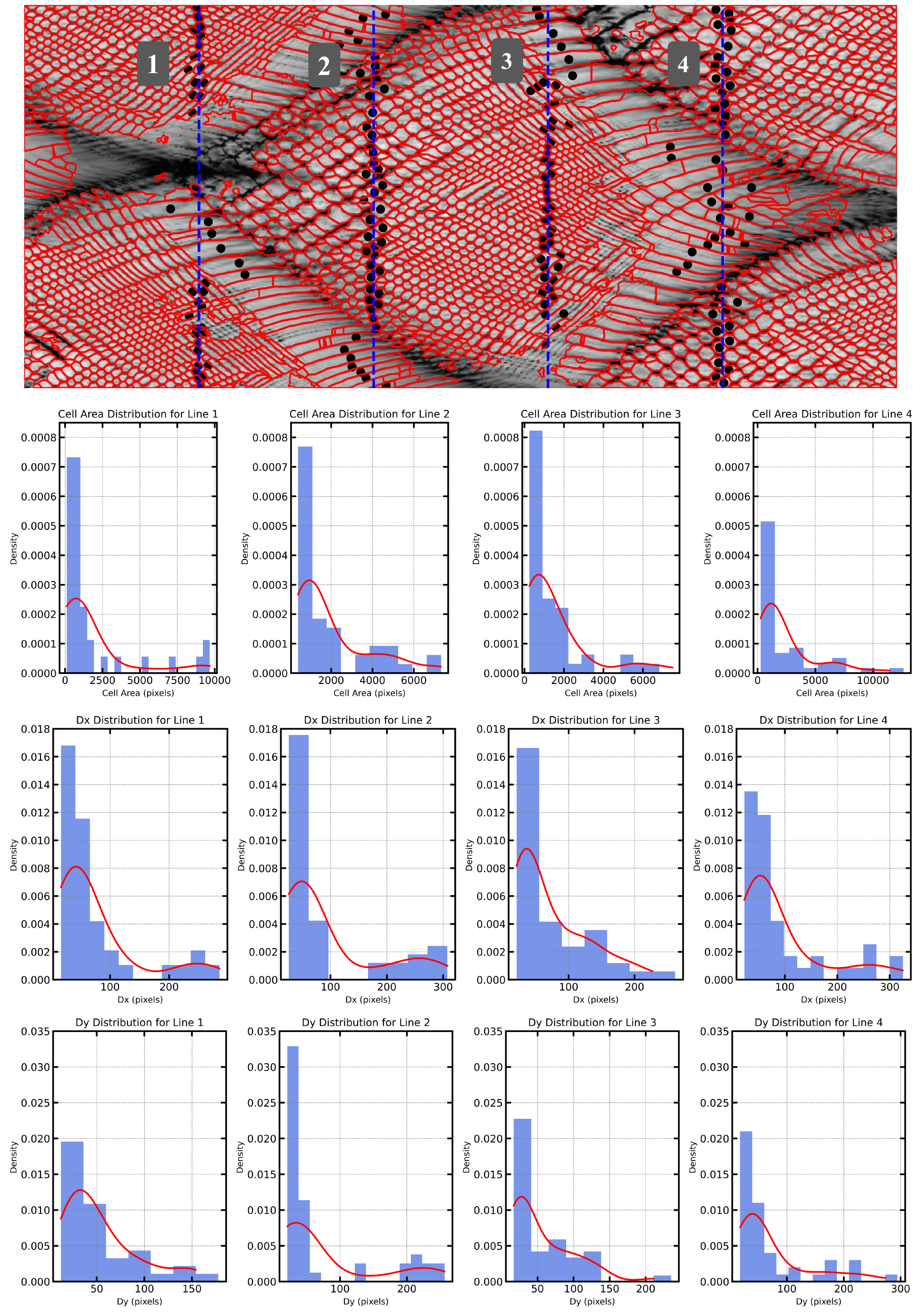}%
    \caption{Statistical analysis of cell characteristics for Fig.\ 2 in \cite{sugiyama2011characteristics}.}
    \label{fig:f2_pdf}
\end{figure}

The soot foil in Fig.~\ref{fig:f2_pdf} presents a distinct and challenging case compared to the previous ones. Here, the soot foil image shows a double cellular pattern, with smaller detonation cells appearing within the broader ones. In this case, the cell area distributions across all four lines strongly skew toward smaller cell sizes, with sharp drop-offs and long tails for larger areas. This skew is more pronounced than in previous images, where the distributions were broader and included more intermediate and larger cell sizes. This illustrates the ability of segmentation to capture a large number of very small cells, a characteristic feature of this particular detonation process. The consistent skew across all four lines indicates similar cellular dynamics across the channel length. However, it is unclear whether the skewed distributions are an inherent trait of the soot foil or whether they result from the difficulty of accurately capturing larger cells within this complex structure. In several regions, densely packed oblong cells with highly skewed dimensions can be observed, which could contribute to the long tails in the distributions. The $Dx$ distributions are highly concentrated towards smaller values, indicating that most cells are closely spaced horizontally. This contrasts sharply with the larger and more variable distributions seen in lines 2, 3, and 4 of the previous soot foil images. The high concentration of low $Dx$ values suggests a denser cellular structure here, but the long tails of the distributions indicate occasional instances of much larger horizontal spacing. This is similarly seen in the $Dy$ distributions, which are heavily skewed towards small values, but occasionally show larger vertical gaps, especially for lines 2, 3, and 4. These long tails may be due to larger gaps in the cellular structure, or possibly mis-segmentation of the image.

The more intricate and disordered cellular structure in Fig.~\ref{fig:f2_pdf} makes it more difficult to obtain accurate cell width measurements. The sharp skew in the cell area and the concentrated tail-heavy distributions in $Dx$ and $Dy$ indicate that the segmentation method is approaching its operational limits. The dense packing of cells, along with occasional larger gaps, reflects a highly complex wave interaction along the detonation front. However, the proposed algorithm can segment the image with reasonable accuracy, highlighting its ability to handle a wide range of cell sizes and orientations.

%%%%%%%%%%%%%%%%%%%%%%%%%%%%%%%%%%%%%%%%%%%%%%%%%%%%%%%%%%%%%%%%%%%%%%%%%%%%%%%
%%%%%%%%%%%%%%%%%%%%%%%%%%%%%%%%%%%%%%%%%%%%%%%%%%%%%%%%%%%%%%%%%%%%%%%%%%%%%%%

\section{Conclusions \label{sec:conclusion}}

An algorithm based on an image segmentation model was used to measure and analyze the morphology of detonation cells. To evaluate the framework, a series of test cases were generated, mimicking soot foil patterns similar to those seen in both experimental and numerical detonation studies. The pretrained ML models demonstrated consistent accuracy, with prediction errors remaining within 10\% even in highly skewed cases. This established the reliability and applicability of the framework. Furthermore, soot foil images from existing literature were analyzed, with histograms and Gaussian KDE fits presented for the areas, horizontal spans ($Dx$), and vertical spans ($Dy$) of the detonation cells. Optimal bin counts for these histograms were computed using a Bayesian technique, ensuring a precise representation of the data.

These statistical analyses were used to quantitatively describe the cellular structures over the channel lengths. For the soot foil in Fig.~\ref{fig:f3_pdf}, the distributions for the cell areas and spans indicated increasing irregularity as the wave progressed. The soot foil in Fig.~\ref{fig:f2_pdf} presented a much more irregular cellular pattern, with the distributions for cell areas and spans being strongly skewed toward small values. However, the long tails of these distributions were indicative of the dense and complex cellular structure with occasionally large, oblong cells. As such, the results demonstrate the algorithm's robustness in capturing statistical trends across diverse datasets, which may include cells with varying sizes and levels of skewness. This illustrates its potential to be broadly utilized in experimental and numerical contexts.

\begin{comment}
Across the three soot foils, clear trends emerge in the distributions of the cell areas and spans. The first soot foil exhibits relatively balanced cell area distributions with moderate variability in $Dx$ and $Dy$, reflecting a more uniform cellular structure. The second soot foil shows increased variability, particularly in cell area and vertical spacing, indicating a shift toward a more irregular and heterogeneous structure. The third soot foil presents the most irregular pattern, with histograms strongly skewed toward smaller cell areas and highly concentrated $Dx$ and $Dy$ distributions. This was indicative of a dense and complex cellular structure with occasionally larger gaps. The progression from the first to the third soot foil corresponds to increasing cellular complexity, and the results demonstrate the algorithm's robustness in capturing statistical trends across diverse datasets. This illustrates its potential to be broadly utilized in experimental and numerical contexts.
\end{comment}

While the algorithm is robust and versatile, the results also highlight the challenges in segmenting and analyzing highly complex or irregular cellular structures. These difficulties can potentially arise from irregular cell shapes and orientations, or imaging issues such as low contrast and noise. Although the current model performs well, its effectiveness could be further enhanced by additional techniques, such as fine-tuning the ML model with a curated dataset to improve detection quality. The potential of the \textit{cyto3} model for three-dimensional analyses is another intriguing possibility. This capability is embedded within the algorithm, although its efficacy and utility remain to be assessed. It is important to note that the detection quality can vary on the basis of the specific characteristics of the soot foil and the image resolution. These factors must be carefully considered to ensure accurate and reliable results.

\section*{Acknowledgments} \addvspace{10pt}

The research is supported by the Air Force Office of Scientific Research (AFOSR) under Grant No.\ FA9550-24-1-0017, with Dr.\ Chiping Li as Program Officer. 
The authors would like to note that while this manuscript was being prepared, another group has independently developed an equivalent approach to determine the size of cells \cite{yokam_communication_wipp}.

\bibliographystyle{elsarticle-num-names.bst} 
\bibliography{bib_cite_abbrev}

\begin{thebibliography}{50}
\expandafter\ifx\csname natexlab\endcsname\relax\def\natexlab#1{#1}\fi
\providecommand{\url}[1]{\texttt{#1}}
\providecommand{\href}[2]{#2}
\providecommand{\path}[1]{#1}
\providecommand{\DOIprefix}{doi:}
\providecommand{\ArXivprefix}{arXiv:}
\providecommand{\URLprefix}{URL: }
\providecommand{\Pubmedprefix}{pmid:}
\providecommand{\doi}[1]{\href{http://dx.doi.org/#1}{\path{#1}}}
\providecommand{\Pubmed}[1]{\href{pmid:#1}{\path{#1}}}
\providecommand{\bibinfo}[2]{#2}
\ifx\xfnm\relax \def\xfnm[#1]{\unskip,\space#1}\fi
%Type = Article
\bibitem[{Raman et~al.(2023)Raman, Prakash, and Gamba}]{raman2023nonidealities}
\bibinfo{author}{V.~Raman}, \bibinfo{author}{S.~Prakash}, \bibinfo{author}{M.~Gamba},
\newblock \bibinfo{title}{Nonidealities in rotating detonation engines},
\newblock \bibinfo{journal}{Annu. Rev. Fluid Mech.} \bibinfo{volume}{55} (\bibinfo{year}{2023}) \bibinfo{pages}{639--674}.
%Type = Article
\bibitem[{Wola{\'n}ski(2013)}]{Wolanski}
\bibinfo{author}{P.~Wola{\'n}ski},
\newblock \bibinfo{title}{Detonative propulsion},
\newblock \bibinfo{journal}{Proc. Combust. Inst.} \bibinfo{volume}{34} (\bibinfo{year}{2013}) \bibinfo{pages}{125--158}.
%Type = Article
\bibitem[{Kailasanath(2000)}]{Kailasanath}
\bibinfo{author}{K.~Kailasanath},
\newblock \bibinfo{title}{Review of propulsion applications of detonation waves},
\newblock \bibinfo{journal}{AIAA J.} \bibinfo{volume}{38} (\bibinfo{year}{2000}) \bibinfo{pages}{1698--1708}.
%Type = Article
\bibitem[{Anand and Gutmark(2019)}]{gutmarkpecs_rde}
\bibinfo{author}{V.~Anand}, \bibinfo{author}{E.~Gutmark},
\newblock \bibinfo{title}{Rotating detonation combustors and their similarities to rocket instabilities},
\newblock \bibinfo{journal}{Prog. Energy Combust. Sci.} \bibinfo{volume}{73} (\bibinfo{year}{2019}) \bibinfo{pages}{182--234}.
%Type = Article
\bibitem[{Yanez et~al.(2015)Yanez, Kuznetsov, and Souto-Iglesias}]{yanez2015analysis}
\bibinfo{author}{J.~Yanez}, \bibinfo{author}{M.~Kuznetsov}, \bibinfo{author}{A.~Souto-Iglesias},
\newblock \bibinfo{title}{An analysis of the hydrogen explosion in the fukushima-daiichi accident},
\newblock \bibinfo{journal}{Int. J. Hydrogen Energy} \bibinfo{volume}{40} (\bibinfo{year}{2015}) \bibinfo{pages}{8261--8280}.
%Type = Article
\bibitem[{Yang et~al.(2021)Yang, Wang, Deng, Dang, Huang, Hu, Li, and Ouyang}]{yang2021review}
\bibinfo{author}{F.~Yang}, \bibinfo{author}{T.~Wang}, \bibinfo{author}{X.~Deng}, \bibinfo{author}{J.~Dang}, \bibinfo{author}{Z.~Huang}, \bibinfo{author}{S.~Hu}, \bibinfo{author}{Y.~Li}, \bibinfo{author}{M.~Ouyang},
\newblock \bibinfo{title}{Review on hydrogen safety issues: Incident statistics, hydrogen diffusion, and detonation process},
\newblock \bibinfo{journal}{Int. J. Hydrogen Energy} \bibinfo{volume}{46} (\bibinfo{year}{2021}) \bibinfo{pages}{31467--31488}.
%Type = Article
\bibitem[{Ng and Lee(2008)}]{ng2008_explosion}
\bibinfo{author}{H.~D. Ng}, \bibinfo{author}{J.~H. Lee},
\newblock \bibinfo{title}{Comments on explosion problems for hydrogen safety},
\newblock \bibinfo{journal}{J. Loss Prev. Process Ind.} \bibinfo{volume}{21} (\bibinfo{year}{2008}) \bibinfo{pages}{136--146}.
%Type = Article
\bibitem[{Short and Quirk(2018)}]{short_arfm_2018}
\bibinfo{author}{M.~Short}, \bibinfo{author}{J.~J. Quirk},
\newblock \bibinfo{title}{High explosive detonation--confiner interactions},
\newblock \bibinfo{journal}{Annu. Rev. Fluid Mech.} \bibinfo{volume}{50} (\bibinfo{year}{2018}) \bibinfo{pages}{215--242}.
%Type = Article
\bibitem[{Voelkel et~al.(2022)Voelkel, Anderson, Short, Chiquete, and Jackson}]{voelkel2022effect}
\bibinfo{author}{S.~J. Voelkel}, \bibinfo{author}{E.~K. Anderson}, \bibinfo{author}{M.~Short}, \bibinfo{author}{C.~Chiquete}, \bibinfo{author}{S.~I. Jackson},
\newblock \bibinfo{title}{Effect of lot microstructure variations on detonation performance of the triaminotrinitrobenzene (tatb)-based insensitive high explosive pbx 9502},
\newblock \bibinfo{journal}{Combust. Flame} \bibinfo{volume}{246} (\bibinfo{year}{2022}) \bibinfo{pages}{112373}.
%Type = Article
\bibitem[{Kuznetsov et~al.(2005)Kuznetsov, Alekseev, Matsukov, and Dorofeev}]{kuznetsov2005ddt}
\bibinfo{author}{M.~Kuznetsov}, \bibinfo{author}{V.~Alekseev}, \bibinfo{author}{I.~Matsukov}, \bibinfo{author}{S.~Dorofeev},
\newblock \bibinfo{title}{Ddt in a smooth tube filled with a hydrogen--oxygen mixture},
\newblock \bibinfo{journal}{Shock Waves} \bibinfo{volume}{14} (\bibinfo{year}{2005}) \bibinfo{pages}{205--215}.
%Type = Article
\bibitem[{Crane et~al.(2019)Crane, Shi, Singh, Tao, and Wang}]{crane2019isolating}
\bibinfo{author}{J.~Crane}, \bibinfo{author}{X.~Shi}, \bibinfo{author}{A.~V. Singh}, \bibinfo{author}{Y.~Tao}, \bibinfo{author}{H.~Wang},
\newblock \bibinfo{title}{Isolating the effect of induction length on detonation structure: Hydrogen--oxygen detonation promoted by ozone},
\newblock \bibinfo{journal}{Combust. Flame} \bibinfo{volume}{200} (\bibinfo{year}{2019}) \bibinfo{pages}{44--52}.
%Type = Article
\bibitem[{Lee(1984)}]{lee1984dynamic}
\bibinfo{author}{J.~H. Lee},
\newblock \bibinfo{title}{Dynamic parameters of gaseous detonations},
\newblock \bibinfo{journal}{Annu. Rev. Fluid Mech.} \bibinfo{volume}{16} (\bibinfo{year}{1984}) \bibinfo{pages}{311--336}.
%Type = Article
\bibitem[{Strehlow(1968)}]{strehlow1968_cf}
\bibinfo{author}{R.~A. Strehlow},
\newblock \bibinfo{title}{Gas pase detonations: recent developments},
\newblock \bibinfo{journal}{Combust. Flame} \bibinfo{volume}{12} (\bibinfo{year}{1968}) \bibinfo{pages}{81--101}.
%Type = Article
\bibitem[{Strehlow and Crooker(1974)}]{strehlow1974_aa}
\bibinfo{author}{R.~A. Strehlow}, \bibinfo{author}{A.~J. Crooker},
\newblock \bibinfo{title}{The structure of marginal detonation waves},
\newblock \bibinfo{journal}{Acta Astronaut.} \bibinfo{volume}{1} (\bibinfo{year}{1974}) \bibinfo{pages}{303--315}.
%Type = Article
\bibitem[{Achasov and Penyazkov(2002)}]{achasov2002dynamics}
\bibinfo{author}{O.~Achasov}, \bibinfo{author}{O.~Penyazkov},
\newblock \bibinfo{title}{Dynamics study of detonation-wave cellular structure 1. statistical properties of detonation wave front},
\newblock \bibinfo{journal}{Shock Waves} \bibinfo{volume}{11} (\bibinfo{year}{2002}) \bibinfo{pages}{297--308}.
%Type = Article
\bibitem[{Pintgen et~al.(2003)Pintgen, Austin, Shepherd, Roy, Frolov, Santoro, and Tsyganov}]{pintgen2003detonation}
\bibinfo{author}{F.~Pintgen}, \bibinfo{author}{J.~Austin}, \bibinfo{author}{J.~Shepherd}, \bibinfo{author}{G.~Roy}, \bibinfo{author}{S.~Frolov}, \bibinfo{author}{R.~Santoro}, \bibinfo{author}{S.~Tsyganov},
\newblock \bibinfo{title}{Detonation front structure: Variety and characterization},
\newblock \bibinfo{journal}{Confined detonations and pulse detonation engines}  (\bibinfo{year}{2003}) \bibinfo{pages}{105--116}.
%Type = Article
\bibitem[{Kellenberger and Ciccarelli(2017)}]{kellenberger2017simultaneous}
\bibinfo{author}{M.~Kellenberger}, \bibinfo{author}{G.~Ciccarelli},
\newblock \bibinfo{title}{Simultaneous schlieren photography and soot foil in the study of detonation phenomena},
\newblock \bibinfo{journal}{Exp. Fluids} \bibinfo{volume}{58} (\bibinfo{year}{2017}) \bibinfo{pages}{1--13}.
%Type = Article
\bibitem[{Crane et~al.(2021)Crane, Shi, Lipkowicz, Kempf, and Wang}]{crane2021geometric}
\bibinfo{author}{J.~Crane}, \bibinfo{author}{X.~Shi}, \bibinfo{author}{J.~T. Lipkowicz}, \bibinfo{author}{A.~M. Kempf}, \bibinfo{author}{H.~Wang},
\newblock \bibinfo{title}{Geometric modeling and analysis of detonation cellular stability},
\newblock \bibinfo{journal}{Proc. Combust. Inst.} \bibinfo{volume}{38} (\bibinfo{year}{2021}) \bibinfo{pages}{3585--3593}.
%Type = Article
\bibitem[{Shepherd(2009)}]{shepherd2009detonation}
\bibinfo{author}{J.~E. Shepherd},
\newblock \bibinfo{title}{Detonation in gases},
\newblock \bibinfo{journal}{Proc. Combust. Inst.} \bibinfo{volume}{32} (\bibinfo{year}{2009}) \bibinfo{pages}{83--98}.
%Type = Article
\bibitem[{Carter and Blunck(2022)}]{carter2022direct}
\bibinfo{author}{M.~Carter}, \bibinfo{author}{D.~L. Blunck},
\newblock \bibinfo{title}{Direct comparison of schlieren and soot foil measurements of detonation cell sizes},
\newblock \bibinfo{journal}{Front. Aerosp. Eng} \bibinfo{volume}{1} (\bibinfo{year}{2022}) \bibinfo{pages}{892330}.
%Type = Book
\bibitem[{Lee(2008)}]{lee_textbook}
\bibinfo{author}{J.~H. Lee}, \bibinfo{title}{The detonation phenomenon}, \bibinfo{publisher}{Cambridge University Press}, \bibinfo{year}{2008}.
%Type = Phdthesis
\bibitem[{Austin(2003)}]{austin_thesis}
\bibinfo{author}{J.~M. Austin}, \bibinfo{title}{The role of instability in gaseous detonation}, Ph.D. thesis, California Institute of Technology, \bibinfo{address}{Pasadena, CA}, \bibinfo{year}{2003}.
%Type = Article
\bibitem[{Jackson and Short(2013)}]{jackson2013_cf}
\bibinfo{author}{S.~I. Jackson}, \bibinfo{author}{M.~Short},
\newblock \bibinfo{title}{The influence of the cellular instability on lead shock evolution in weakly unstable detonation},
\newblock \bibinfo{journal}{Combust. Flame} \bibinfo{volume}{160} (\bibinfo{year}{2013}) \bibinfo{pages}{2260--2274}.
%Type = Article
\bibitem[{Ciccarelli et~al.(1994)Ciccarelli, Ginsberg, Boccio, Economos, Sato, and Kinoshita}]{ciccarelli1994detonation}
\bibinfo{author}{G.~Ciccarelli}, \bibinfo{author}{T.~Ginsberg}, \bibinfo{author}{J.~Boccio}, \bibinfo{author}{C.~Economos}, \bibinfo{author}{K.~Sato}, \bibinfo{author}{M.~Kinoshita},
\newblock \bibinfo{title}{Detonation cell size measurements and predictions in hydrogen-air-steam mixtures at elevated temperatures},
\newblock \bibinfo{journal}{Combust. Flame} \bibinfo{volume}{99} (\bibinfo{year}{1994}) \bibinfo{pages}{212--220}.
%Type = Techreport
\bibitem[{Stamps et~al.(1991)Stamps, Benedick, and Tieszen}]{stamps1991hydrogen}
\bibinfo{author}{D.~Stamps}, \bibinfo{author}{W.~B. Benedick}, \bibinfo{author}{S.~Tieszen}, \bibinfo{title}{Hydrogen-air-diluent detonation study for nuclear reactor safety analyses}, \bibinfo{type}{Technical Report}, US Nuclear Regulatory Commission, \bibinfo{year}{1991}.
%Type = Article
\bibitem[{Vasil'ev(2006)}]{vasil2006cell}
\bibinfo{author}{A.~Vasil'ev},
\newblock \bibinfo{title}{Cell size as the main geometric parameter of a multifront detonation wave},
\newblock \bibinfo{journal}{J. Propuls. Power} \bibinfo{volume}{22} (\bibinfo{year}{2006}) \bibinfo{pages}{1245--1260}.
%Type = Article
\bibitem[{Meagher et~al.(2023)Meagher, Shi, Santos, Muraleedharan, Crane, Poludnenko, Wang, and Zhao}]{alexei2023}
\bibinfo{author}{P.~A. Meagher}, \bibinfo{author}{X.~Shi}, \bibinfo{author}{J.~P. Santos}, \bibinfo{author}{N.~K. Muraleedharan}, \bibinfo{author}{J.~Crane}, \bibinfo{author}{A.~Y. Poludnenko}, \bibinfo{author}{H.~Wang}, \bibinfo{author}{X.~Zhao},
\newblock \bibinfo{title}{Isolating gasdynamic and chemical effects on the detonation cellular structure: A combined experimental and computational study},
\newblock \bibinfo{journal}{Proc. Combust. Inst.} \bibinfo{volume}{39} (\bibinfo{year}{2023}) \bibinfo{pages}{2865--2873}.
%Type = Inproceedings
\bibitem[{Prakash et~al.(2019)Prakash, Fi{\'e}vet, and Raman}]{prakash2019_fuelstrat}
\bibinfo{author}{S.~Prakash}, \bibinfo{author}{R.~Fi{\'e}vet}, \bibinfo{author}{V.~Raman},
\newblock \bibinfo{title}{The effect of fuel stratification on the detonation wave structure},
\newblock  \bibinfo{booktitle}{AIAA SciTech 2019 Forum}, \bibinfo{year}{2019}, p. \bibinfo{pages}{1511}.
%Type = Article
\bibitem[{Kumar(1990)}]{kumar1990detonation}
\bibinfo{author}{R.~Kumar},
\newblock \bibinfo{title}{Detonation cell widths in hydrogen oxygen diluent mixtures},
\newblock \bibinfo{journal}{Combust. Flame} \bibinfo{volume}{80} (\bibinfo{year}{1990}) \bibinfo{pages}{157--169}.
%Type = Article
\bibitem[{Ullman et~al.(2024)Ullman, Prakash, Barwey, and Raman}]{ullman-2024}
\bibinfo{author}{M.~Ullman}, \bibinfo{author}{S.~Prakash}, \bibinfo{author}{S.~Barwey}, \bibinfo{author}{V.~Raman},
\newblock \bibinfo{title}{Detonation structure in the presence of mixture stratification using reaction-resolved simulations},
\newblock \bibinfo{journal}{Combust. Flame} \bibinfo{volume}{264} (\bibinfo{year}{2024}) \bibinfo{pages}{113427}.
%Type = Inproceedings
\bibitem[{Shepherd et~al.(1988)Shepherd, Moen, Murray, and Thibault}]{shepherd1988analyses}
\bibinfo{author}{J.~Shepherd}, \bibinfo{author}{I.~Moen}, \bibinfo{author}{S.~Murray}, \bibinfo{author}{P.~Thibault},
\newblock \bibinfo{title}{Analyses of the cellular structure of detonations},
\newblock  \bibinfo{booktitle}{Symp. Combust. Proc.}, volume~\bibinfo{volume}{21}, \bibinfo{organization}{Elsevier}, \bibinfo{year}{1988}, pp. \bibinfo{pages}{1649--1658}.
%Type = Article
\bibitem[{Nair et~al.(2023)Nair, Keller, Minesi, Pineda, and Spearrin}]{nair2023detonation}
\bibinfo{author}{A.~P. Nair}, \bibinfo{author}{A.~R. Keller}, \bibinfo{author}{N.~Q. Minesi}, \bibinfo{author}{D.~I. Pineda}, \bibinfo{author}{R.~M. Spearrin},
\newblock \bibinfo{title}{Detonation cell size of liquid hypergolic propellants: Estimation from a non-premixed combustor},
\newblock \bibinfo{journal}{Proc. Combust. Inst.} \bibinfo{volume}{39} (\bibinfo{year}{2023}) \bibinfo{pages}{2757--2765}.
%Type = Article
\bibitem[{Ng et~al.(2024)Ng, Hoffman, Pineda, and Combs}]{ng2024detonation}
\bibinfo{author}{B.~M. Ng}, \bibinfo{author}{E.~N. Hoffman}, \bibinfo{author}{D.~I. Pineda}, \bibinfo{author}{C.~S. Combs},
\newblock \bibinfo{title}{Detonation cell size estimation via chemiluminescence imaging in an optically accessible linear detonation tube},
\newblock \bibinfo{journal}{Exp. Fluids} \bibinfo{volume}{65} (\bibinfo{year}{2024}) \bibinfo{pages}{108}.
%Type = Article
\bibitem[{Siatkowski et~al.(2023)Siatkowski, Wacko, and Kindracki}]{siatkowski2021experimental}
\bibinfo{author}{S.~Siatkowski}, \bibinfo{author}{K.~Wacko}, \bibinfo{author}{J.~Kindracki},
\newblock \bibinfo{title}{Extensive study on the detonation cell size of biogas-oxygen mixtures},
\newblock \bibinfo{journal}{Fuel} \bibinfo{volume}{344} (\bibinfo{year}{2023}) \bibinfo{pages}{128016}.
%Type = Article
\bibitem[{Siatkowski et~al.(2024)Siatkowski, Wacko, and Kindracki}]{siatkowski2024predicting}
\bibinfo{author}{S.~Siatkowski}, \bibinfo{author}{K.~Wacko}, \bibinfo{author}{J.~Kindracki},
\newblock \bibinfo{title}{Predicting detonation cell size of biogas--oxygen mixtures using machine learning models},
\newblock \bibinfo{journal}{Shock Waves}  (\bibinfo{year}{2024}) \bibinfo{pages}{1--9}.
%Type = Article
\bibitem[{Bakalis et~al.(2023)Bakalis, Valipour, Bentahar, Kadem, Teng, and Ng}]{bakalis2023detonation}
\bibinfo{author}{G.~Bakalis}, \bibinfo{author}{M.~Valipour}, \bibinfo{author}{J.~Bentahar}, \bibinfo{author}{L.~Kadem}, \bibinfo{author}{H.~Teng}, \bibinfo{author}{H.~D. Ng},
\newblock \bibinfo{title}{Detonation cell size prediction based on artificial neural networks with chemical kinetics and thermodynamic parameters},
\newblock \bibinfo{journal}{Fuel Comm.} \bibinfo{volume}{14} (\bibinfo{year}{2023}) \bibinfo{pages}{100084}.
%Type = Techreport
\bibitem[{Kaneshige and Shepherd(1999)}]{Shepherd-database}
\bibinfo{author}{M.~Kaneshige}, \bibinfo{author}{J.~Shepherd}, \bibinfo{title}{Detonation database}, \bibinfo{type}{Technical Report}, California Institute of Technology, \bibinfo{year}{1999}. \bibinfo{note}{{GALCIT Technical Report FM97-8}}.
%Type = Article
\bibitem[{Suri(2000)}]{suri-2000}
\bibinfo{author}{J.~S. Suri},
\newblock \bibinfo{title}{{Computer Vision, pattern recognition and image processing in left ventricle segmentation: the last 50 years}},
\newblock \bibinfo{journal}{Pattern Anal. Appl.} \bibinfo{volume}{3} (\bibinfo{year}{2000}) \bibinfo{pages}{209--242}.
%Type = Article
\bibitem[{Stringer et~al.(2020)Stringer, Wang, Michaelos, and Pachitariu}]{cellpose_stringer-2020}
\bibinfo{author}{C.~Stringer}, \bibinfo{author}{T.~Wang}, \bibinfo{author}{M.~Michaelos}, \bibinfo{author}{M.~Pachitariu},
\newblock \bibinfo{title}{{Cellpose: a generalist algorithm for cellular segmentation}},
\newblock \bibinfo{journal}{Nat. Methods} \bibinfo{volume}{18} (\bibinfo{year}{2020}) \bibinfo{pages}{100--106}.
%Type = Inproceedings
\bibitem[{Ronneberger et~al.(2015)Ronneberger, Fischer, and Brox}]{UNet}
\bibinfo{author}{O.~Ronneberger}, \bibinfo{author}{P.~Fischer}, \bibinfo{author}{T.~Brox},
\newblock \bibinfo{title}{U-net: Convolutional networks for biomedical image segmentation},
\newblock  \bibinfo{booktitle}{Medical image computing and computer-assisted intervention--MICCAI 2015: 18th international conference, Munich, Germany, October 5-9, 2015, proceedings, part III 18}, \bibinfo{organization}{Springer}, \bibinfo{year}{2015}, pp. \bibinfo{pages}{234--241}.
%Type = Inproceedings
\bibitem[{Li and Wand(2016)}]{styletransfer}
\bibinfo{author}{C.~Li}, \bibinfo{author}{M.~Wand},
\newblock \bibinfo{title}{Precomputed real-time texture synthesis with markovian generative adversarial networks},
\newblock  \bibinfo{booktitle}{Computer Vision--ECCV 2016: 14th European Conference, Amsterdam, The Netherlands, October 11-14, 2016, Proceedings, Part III 14}, \bibinfo{organization}{Springer}, \bibinfo{year}{2016}, pp. \bibinfo{pages}{702--716}.
%Type = Article
\bibitem[{Zargari et~al.(2024)Zargari, Topacio, Mashhadi, and Shariati}]{GANcellpose_zargari-2024}
\bibinfo{author}{A.~Zargari}, \bibinfo{author}{B.~R. Topacio}, \bibinfo{author}{N.~Mashhadi}, \bibinfo{author}{S.~A. Shariati},
\newblock \bibinfo{title}{{Enhanced Cell Segmentation with Limited Training Datasets using Cycle Generative Adversarial Networks}},
\newblock \bibinfo{journal}{iScience} \bibinfo{volume}{27} (\bibinfo{year}{2024}) \bibinfo{pages}{109740}.
%Type = Article
\bibitem[{Stringer and Pachitariu(2024)}]{Cellpose3}
\bibinfo{author}{C.~Stringer}, \bibinfo{author}{M.~Pachitariu},
\newblock \bibinfo{title}{Cellpose3: one-click image restoration for improved cellular segmentation},
\newblock \bibinfo{journal}{bioRxiv}  (\bibinfo{year}{2024}) \bibinfo{pages}{2024--02}.
%Type = Article
\bibitem[{Smirnov et~al.(2024)Smirnov, Nikitin, Mikhalchenko, Stamov, and Tyurenkova}]{smirnov2024modelling}
\bibinfo{author}{N.~Smirnov}, \bibinfo{author}{V.~Nikitin}, \bibinfo{author}{E.~Mikhalchenko}, \bibinfo{author}{L.~Stamov}, \bibinfo{author}{V.~Tyurenkova},
\newblock \bibinfo{title}{Modelling cellular structure of detonation waves in hydrogen-air mixtures},
\newblock \bibinfo{journal}{Int. J. Hydrogen Energy} \bibinfo{volume}{49} (\bibinfo{year}{2024}) \bibinfo{pages}{495--509}.
%Type = Article
\bibitem[{Li et~al.(2021)Li, Han, Li, and Fan}]{li2021influences}
\bibinfo{author}{H.~Li}, \bibinfo{author}{W.~Han}, \bibinfo{author}{J.~Li}, \bibinfo{author}{W.~Fan},
\newblock \bibinfo{title}{Influences of incoming flow on re-initiation of cellular detonations},
\newblock \bibinfo{journal}{Combust. Flame} \bibinfo{volume}{229} (\bibinfo{year}{2021}) \bibinfo{pages}{111376}.
%Type = Article
\bibitem[{Sugiyama and Matsuo(2011)}]{sugiyama2011characteristics}
\bibinfo{author}{Y.~Sugiyama}, \bibinfo{author}{A.~Matsuo},
\newblock \bibinfo{title}{On the characteristics of two-dimensional double cellular detonations with two successive reactions model},
\newblock \bibinfo{journal}{Proc. Combust. Inst.} \bibinfo{volume}{33} (\bibinfo{year}{2011}) \bibinfo{pages}{2227--2233}.
%Type = Article
\bibitem[{Sharma and Raman(2024)}]{vanshRAG}
\bibinfo{author}{V.~Sharma}, \bibinfo{author}{V.~Raman},
\newblock \bibinfo{title}{A reliable knowledge processing framework for combustion science using foundation models},
\newblock \bibinfo{journal}{Energy AI} \bibinfo{volume}{16} (\bibinfo{year}{2024}) \bibinfo{pages}{100365}.
%Type = Article
\bibitem[{Knuth(2006)}]{knuth2006optimal}
\bibinfo{author}{K.~H. Knuth},
\newblock \bibinfo{title}{Optimal data-based binning for histograms},
\newblock \bibinfo{journal}{arXiv preprint physics/0605197}  (\bibinfo{year}{2006}).
%Type = Misc
\bibitem[{Kozak(2024)}]{yokam_communication_wipp}
\bibinfo{author}{Y.~Kozak}, \bibinfo{title}{{Personal communication during WiPP session at 40th Internation Symposium on Combustion, Milan, Italy}}, \bibinfo{year}{2024}.
%Type = Inproceedings
\bibitem[{Monnier et~al.(2022)Monnier, Rodriguez, Vidal, and Zitoun}]{monnier-icders}
\bibinfo{author}{V.~Monnier}, \bibinfo{author}{V.~Rodriguez}, \bibinfo{author}{P.~Vidal}, \bibinfo{author}{R.~Zitoun},
\newblock \bibinfo{title}{{Experimental analysis of cellular detonations: a discussion on regularity and three-dimensional patterns}},
\newblock  \bibinfo{booktitle}{{28th International Colloquium on the Dynamics of Explosions and Reactive Systems}}, \bibinfo{address}{Naples, Italy}, \bibinfo{year}{2022}.

\end{thebibliography}

%%%%%%%%%%%%%%%%%%%%%%%%%%%%%%%%%%%%%%%%%%%%%%%%%%%%%%%%%%%%%%%%%%%%%%%%%%%%%%%
%%%%%%%%%%%%%%%%%%%%%%%%%%%%%%%%%%%%%%%%%%%%%%%%%%%%%%%%%%%%%%%%%%%%%%%%%%%%%%%
\clearpage

\appendix
\section{Effect of data preprocessing}\label{appendix:preProc}

\begin{figure}[!hbt]
    \centering
    \includegraphics[clip,width=0.95\textwidth]{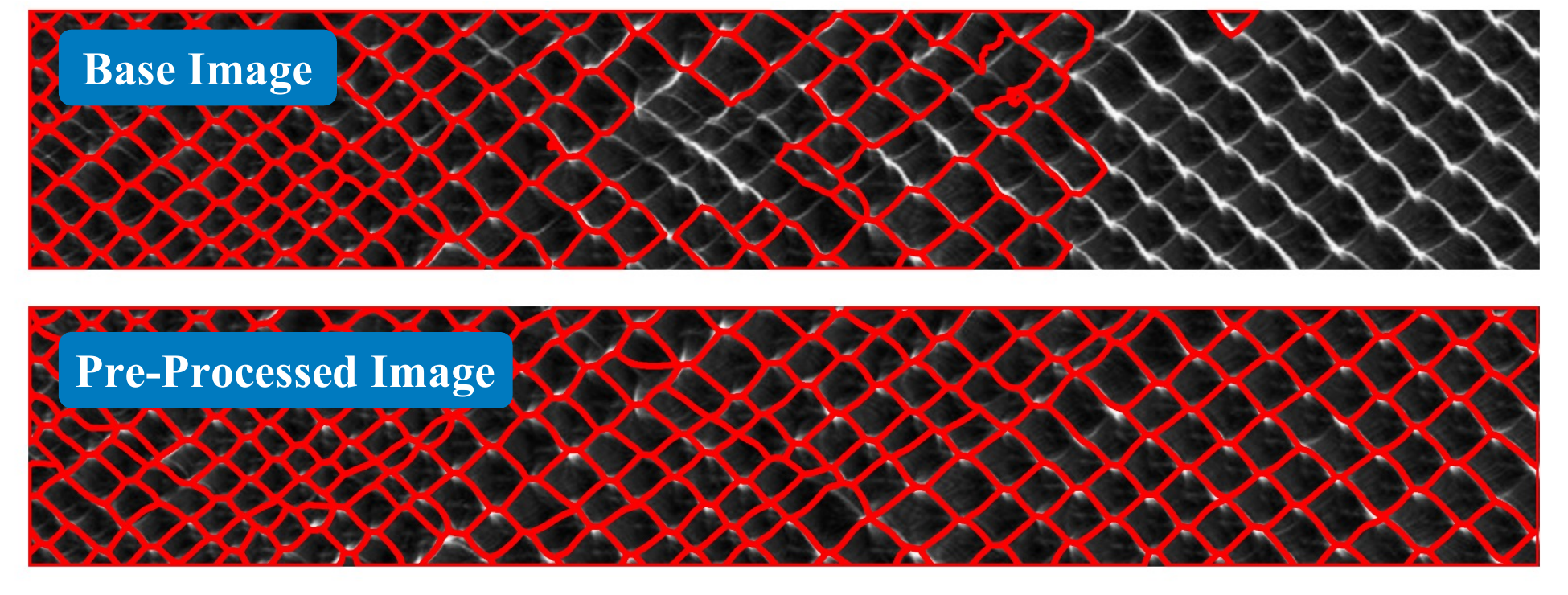}%
    \caption{Detected cells marked with red outlines overlaid on figures from Fig.\ 10 in \cite{smirnov2024modelling} for segmentation using base image and a preprocessed image.}
    \label{fig:smi_f10a}
\end{figure}

The preprocessing steps outlined in Section \ref{sec:preProc} are optional and depend on the quality of the data. However, they can play a crucial role in preparing the data to be input into the ML model. Figure \ref{fig:smi_f10a} demonstrates the impacts of these operations. The top row image, which lacked preprocessing, shows unsuccessful cell segmentation. Meanwhile, the bottom row image, after preprocessing, achieves successful segmentation. Here, preprocessing consists of first applying directional gradients in the north and south directions, followed by line dilation. Specifically, two dilation operations are performed on the north-directed lines, which are thinner in this image. Then, the arrays are combined into a single input array. Proper data preparation is vital for the accuracy of any algorithm, including the one proposed in this study. Although the sequence and number of iterations for each operation may vary depending on the specific case, the key is identifying which features require refinement. This understanding is essential for optimizing the algorithm's performance.

\section{Additional data analysis}\label{appendix:extraData}
The following images are extracted from articles focused on experimental analysis, demonstrating the model's ability to directly segment soot foils from experimental records. 

\begin{figure}[!hbt]
    \centering
    \includegraphics[clip,width=0.95\textwidth]{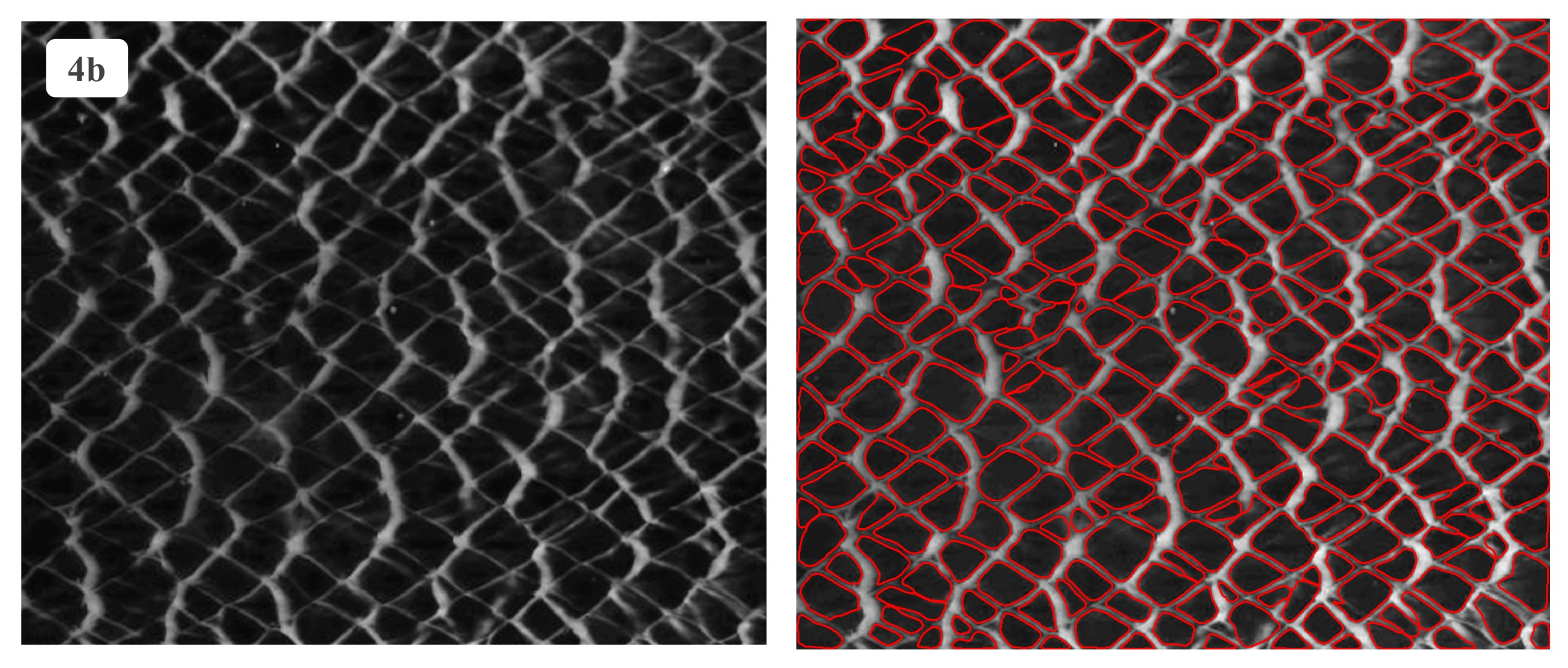}%
    \caption{Detected cells marked with red outlines overlaid on soot foil from Fig.\ 4 in \cite{crane2019isolating}.}
    \label{fig:haiw4b}
\end{figure}

\begin{figure}[!hbt]
    \centering
    \includegraphics[clip,width=0.95\textwidth]{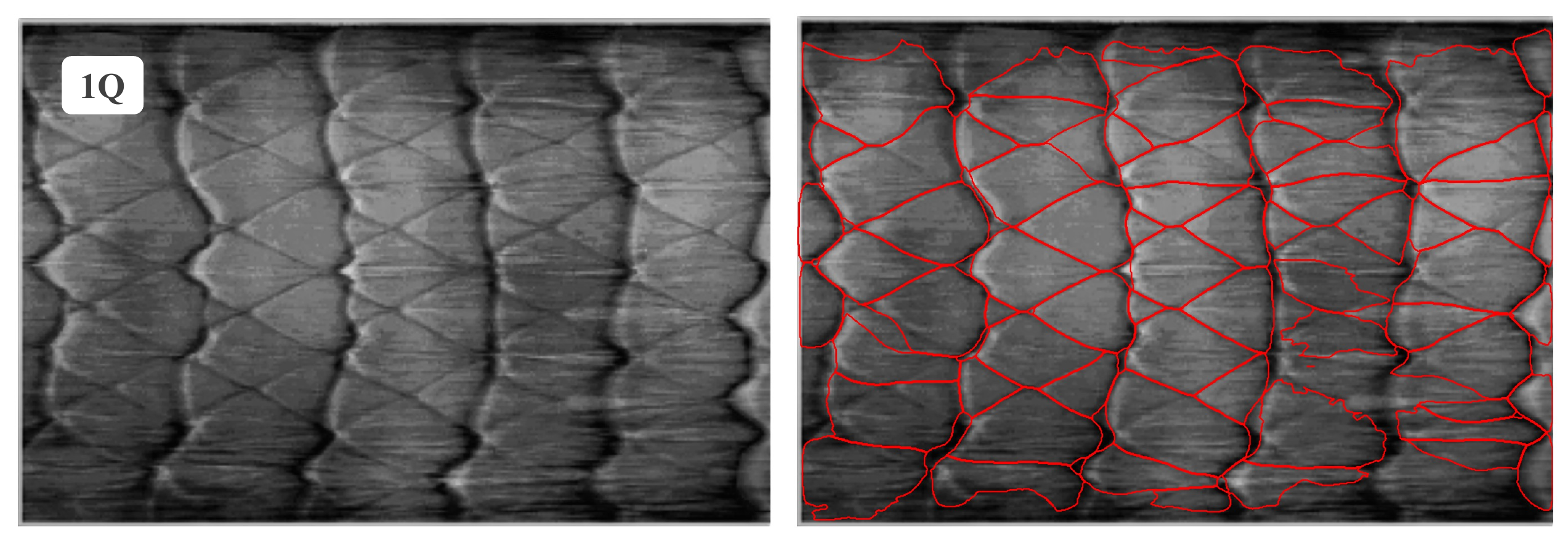}%
    \caption{Detected cells marked with red outlines overlaid on soot foil from Fig.\ 1 in \cite{monnier-icders}.}
    \label{fig:icders1q}
\end{figure}

\end{document}